\pdfoutput=1
\documentclass{article}

\PassOptionsToPackage{numbers,compress}{natbib}
\usepackage[preprint]{neurips_2026}

\usepackage[utf8]{inputenc} % allow utf-8 input
\usepackage[T1]{fontenc}    % use 8-bit T1 fonts
\usepackage{hyperref}       % hyperlinks
\usepackage{url}            % simple URL typesetting
\usepackage{booktabs}       % professional-quality tables
\usepackage{amsfonts}       % blackboard math symbols
\usepackage{nicefrac}       % compact symbols for 1/2, etc.
\usepackage{microtype}      % microtypography
\usepackage{xcolor}         % colors
\usepackage{graphicx}
\usepackage{subcaption}
\usepackage{array}
\usepackage{booktabs}
\usepackage{array}
\usepackage{float}
\usepackage{pifont}
\usepackage{xcolor}
\usepackage{enumitem}
\usepackage{placeins}
\usepackage{amsmath}
\usepackage{multirow}
\usepackage{xcolor}
\usepackage{colortbl}
\usepackage{amsfonts}
\usepackage{wrapfig}
\usepackage[dvipsnames]{xcolor}
\usepackage{booktabs}
\usepackage{multirow}
\usepackage[dvipsnames]{xcolor}
\usepackage{makecell}

\newcommand{\E}{\mathbb{E}}

\newcommand{\clip}{\operatorname{clip}}
\newcommand{\Shift}{\operatorname{Shift}}
\newcommand{\WDisp}{\operatorname{WDisp}}
\newcommand{\refmodel}{\mathrm{ref}}

 \definecolor{darkgreen}{RGB}{0,128,0}
 \usepackage[x11names,dvipsnames]{xcolor}         % colors
\usepackage{makecell} % For multi-line cells
\title{GRPO Training Dynamics for Small Language Models}

\author{%
\textbf{
Rajat Ghosh$^{1,*}$,
Vaishnavi Bhargava$^{1}$,
Henry Wong$^{1}$,
Aryan Singhal$^{1}$,
Debojyoti Dutta$^{1}$
}
\\[0.5em]
$^{1}$Nutanix
\\[0.5em]
$^{*}$Correspondence: \texttt{rajat.ghosh11@gmail.com}
}

\begin{document}

\maketitle

\begin{abstract}

Group Relative Policy Optimization (GRPO) has emerged as a memory-efficient reinforcement fine-tuning (RFT) technique for reasoning-intensive tasks. However, GRPO training dynamics on small language models (SLMs) remain poorly understood, limiting its reliable adoption and reproducibility in open and resource-constrained environments. In this work, we present a systematic study of GRPO fine-tuning for SLMs ranging from 1.5B to 7B parameters under a practical single-node 8×A100 compute budget. Our study spans multiple model families and reasoning domains, including mathematics, coding, and multiple-choice question answering (MCQ) in science. Across these settings, we analyze how group size affects policy convergence, training stability, and downstream benchmark performance. We further characterize tensor-level update dynamics during GRPO training and investigate whether the choice of LoRA target modules and layers can improve the performance of GRPO-tuned models. While our initial GRPO-tuned models outperform their base counterparts on approximately 80\% of mathematical benchmark evaluations, they demonstrate limited capability on MCQ and code reasoning tasks. Guided by our mechanistic evaluations, we refined our LoRA and reward-shaping configurations to improve performance in latter domains. These findings provide practical guidance for GRPO training for SLMs.

\end{abstract}

\section{Introduction}

% Large language models have demonstrated significant capabilities in complex problem-solving, yet aligning them for reasoning-intensive tasks requires robust post-training. Traditional reinforcement learning approaches, such as Proximal Policy Optimization (PPO), rely on an auxiliary learned value model to estimate advantages, which imposes substantial memory and computational overhead. To alleviate this, 

Group Relative Policy Optimization (GRPO) has emerged as an efficient RLHF technique \citep{shao2024deepseekmath, guo2025deepseek}. By replacing the separate value function with a sample-based advantage mechanism, GRPO strictly reduces memory requirements, making advanced reasoning alignment viable for resource-constrained environments. Crucially, because it relies on comparing actual outcomes rather than a learned critic's estimations, GRPO inherently requires tasks with objective, verifiable rewards. This makes it an ideal framework for reasoning domains such as mathematics and code generation. GRPO has emerged as a promising alternative due to its computational efficiency. Open-source efforts such as Open-R1/TRL \citep{openr1}, HybridFlow/Verl \citep{sheng2025hybridflow}, and NVIDIA-NeMo/RL \citep{nemo-rl} have accelerated experimentation with GRPO. Despite this growing ecosystem, the training dynamics of GRPO remain poorly understood, especially for SLMs operating under tight compute budgets. In contrast to PPO and related methods, there is limited empirical and theoretical analysis characterizing convergence behavior, reward saturation, or compute–performance trade-offs. Moreover, the mechanistic interpretability continues to be a critical gap in deep learning research.

Agentic and edge AI systems increasingly rely on SLMs \citep{belcak2025small} with fewer than 8B parameters to reduce latency, lower energy consumption, and minimize operational costs \citep{liu2024mobilellm, abdin2024phi}. As interest in SLMs grows, memory-efficient reinforcement learning fine-tuning methods such as GRPO and GRPO-LoRA \cite{wang2025tina, hu2021loralowrankadaptationlarge} are becoming an increasingly important research direction for improving reasoning under constrained resources.  However, SLMs face unique challenges during this post-training phase. Due to a restricted parametric surface area, SLMs lack the representational bandwidth required to adapt to complex downstream reasoning tasks while maintaining their pre-trained general capabilities \citep{hadjikyriacou2024would, luo2025through}. Under the strong optimization pressure of GRPO, maintaining a diverse output distribution becomes inefficient for the policy; consequently, the distribution sharply narrows, discarding pre-trained behaviors to over-optimize for a single, high-reward output pattern \citep{ziegler2019fine, mohammadi2024creativity}. This dynamic exposes SLM GRPO training to severe failure modes, primarily mode collapse, catastrophic forgetting, and naive reward hacking \citep{gao2023scaling, rafailov2024scaling}. Unlike standard supervised fine-tuning, hyperparameter optimization and performance tuning in GRPO present a greater formidable challenge in absence of value functions due to the non-stationary dynamics of on-policy rollouts, reward sparsity, and the delicate equilibrium required between reward maximization and KL-regularization \cite{wang2026treestylebranchingmattersthought, zhang2025gvpogroupvariancepolicy}. Traditional hyperparameter optimization relies predominantly on behavioral metrics and validation loss, which can inadvertently select for models that rely on brittle heuristics rather than robust internal representations. To address this, recent approaches advocate for performance improvement through hyperparameter tuning guided by mechanistic evaluation. By leveraging tools such as causal interventions and monitoring the emergence of specific induction circuits during training, practitioners can move beyond black-box optimization \cite{arora2026mechanisticevaluationtransformersstate, sun2025hyperdasautomatingmechanisticinterpretability}. This mechanistic feedback provides a high-resolution signal, ensuring that hyperparameter configurations such as learning rate schedules, and sparsity constraints explicitly drive the model to learn generalizable, algorithmically sound mechanisms rather than surface-level statistical mimicry. Incorporation of mechanistic feedback for improving GRPO fine-tuning is largely missing from 
GRPO fine-tuning especially in resource-constraint environments \cite{pikus2025hardexamplesneedmaximizing}. Addressing these challenges, the main contributions of this paper are as follows:

\begin{itemize}[leftmargin=*]
\item \textbf{Empirical characterization of GRPO dynamics.} We provide a comprehensive analysis of training dynamics across three distinct SLM families trained on three diverse reasoning datasets: GSM8K (mathematical reasoning), ARC-Challenge (closed-book QA), and OpenCoder (code generation).
\item \textbf{Performance improvement guided by mechanistic evaluation.} Informed by mechanistic evaluations, we tune hyper-parameters to enable targeted GRPO performance improvement. 
\item \textbf{Reproducible pipeline.} We open-source our GRPO post-training pipeline.
\end{itemize}

\section{Preliminaries}

\label{sec:preliminaries}

% Group Relative Policy Optimization (GRPO) \cite{guo2025deepseek,shao2024deepseekmath} is an approximate policy-iteration method that eliminates the need for a separate value function (critic) by estimating baselines directly from group samples.  Prior work has shown that critic-based methods, such as PPO, use learned value functions to reduce gradient variance and stabilize optimization \cite{schulman2017ppo, openai_spinningup_pg}, whereas GRPO training, in the absence of value functions, may become more sensitive to reward scaling, clipping, KL regularization, and group size. 

Group Relative Policy Optimization (GRPO) 
\cite{guo2025deepseek,shao2024deepseekmath} is an approximate policy optimization method that replaces the learned critic with intra-group baselines estimated directly from sampled rollouts for each prompt. Critic-based methods such as PPO use learned value functions to reduce gradient variance and stabilize training \cite{schulman2017ppo, openai_spinningup_pg}; without this stabilizer, GRPO can be more sensitive to the choice of reward scaling, clipping threshold, KL regularization coefficient, and group size.

We formulate GRPO training design challenge as a bilevel optimization problem. The inner problem performs GRPO training for a fixed design parameter space \(\lambda\), while the outer problem refines \(\lambda\) to maximize target performance subject to behavioral, output-level, activation-level, and weight-level constraints. This outer optimization problem (also known as hyperparameter tuning) can use randomized grid search \cite{Florea_2019} and Bayesian search \cite{snoek2012practicalbayesianoptimizationmachine}. However, GRPO is notably more complex than standard supervised learning, requiring a precise calibration of reward signals, group-based advantage estimators, and KL constraints to stabilize the non-stationary training dynamics. The GRPO training design parameter space includes different training hyper-parameters, data curation parameters, and reward shaping weights, as shown below:

\begin{equation}
    \lambda
    =
    \left(
    \eta,
    \epsilon,
    \beta,
    G,
    B,
    W_{\mathrm{LoRA}},
    L_{\mathrm{LoRA}},
    r_{\mathrm{LoRA}},
    \alpha_{\mathrm{LoRA}},
    \mathcal{D}_{\mathrm{mix}},
    w_R
    \right)
\end{equation}
Here \(\eta\) is the learning rate, \(\epsilon\) is
the clipping threshold, \(\beta\) is the KL coefficient, \(G\) is the GRPO group
size, \(B\) is the batch size, \(W_{\mathrm{LoRA}}\) is LoRA modules, \(L_{\mathrm{LoRA}}\) is LoRA layers,
\(r_{\mathrm{LoRA}}\) is LoRA rank,  \(\alpha_{\mathrm{LoRA}}\) is scaling factor, \(\mathcal{D}_{\mathrm{mix}}\) is the data mixture, and
\(w_R\) denotes reward weights. The outer optimization problem of searching an optimal parameter space, $\lambda$, can use traditional black-box methods. However, these methods could be cost-prohibitive. An alternative approach could be mechanistic evaluation-guided tuning. The metrics, including weight displacement and activation shift, can be used to diagnose training behavior and guide targeted hyperparameter refinement. A full discussion of these diagnostics is presented in \ref{app:mech-feedback}.

Formally, for each query $q \sim P(Q)$, GRPO samples $G$ outputs
$\{o_1, \ldots, o_G\} \sim \pi_{\theta_{\mathrm{old}}}(O \mid q)$
with rewards $\{r_1, \ldots, r_G\}$. The group-relative advantage
$A_i = (r_i - \bar{r}) / \sigma_r$, where $\bar{r}$ and $\sigma_r$
denote the intra-group reward mean and standard deviation, normalizes
each output relative to its group. The policy $\pi_\theta$ is updated
by maximizing:
\begin{equation}
\mathcal{J}_{\mathrm{GRPO}}(\theta)
= \mathbb{E}_{q,\{o_i\}}\!\left[
\tfrac{1}{G}\textstyle\sum_{i=1}^{G}
\min\!\Big(\rho_i A_i,\;
\mathrm{clip}(\rho_i,\, 1{-}\epsilon,\, 1{+}\epsilon)\, A_i\Big)
- \beta\, D_{\mathrm{KL}}(\pi_\theta \,\Vert\, \pi_{\mathrm{ref}})
\right],
\label{eq:grpo-objective}
\end{equation}
where $\rho_i = \pi_\theta(o_i \mid q) /
\pi_{\theta_{\mathrm{old}}}(o_i \mid q)$ is the importance sampling
ratio and $\pi_{\mathrm{ref}}$ is a fixed reference policy. A complete
formulation is provided in Appendix~\ref{app:grpo-equation}.

\section{Experimental design}

This work is situated at the intersection of the empirical characterization of GRPO training dynamics and mechanistic interpretability. In consonance with that, we pose the following research questions: 

\begin{itemize}[leftmargin=*]
    \item \textbf{RQ1.} Does GRPO-only post-training improve reasoning capabilities for small language models?
    \item \textbf{RQ2.} Can we use mechanistic feedback to improve training performance?
\end{itemize}

To investigate these research questions and ensure empirical robustness, we select three SLMs spanning diverse parameter scales, architectures, and training paradigms: DeepSeek-R1-Distill-Qwen-1.5B (hereafter, DeepSeek-R1-Qwen-1.5B) \cite{DeepSeek-R1-Qwen-1.5B}, Nemotron-Mini-4B-Instruct (hereafter, Nemotron-4B) \cite{Nemotron-4B}, and DeepSeek-LLM-7B-Chat (hereafter, DeepSeek-7B) \cite{DeepSeek-7B}. To maximize the generalizability of our analysis across distinct cognitive domains, we evaluate these models on three specialized reasoning datasets: GSM8K \cite{cobbe2021gsm8k} for mathematical problem-solving, ARC-Challenge (hereafter, ARC) for closed-book scientific reasoning, and the OpenCoder educational instruction split (hereafter, OpenCoder) for code generation. Across all model-dataset combinations, we systematically analyze the sensitivity of GRPO training dynamics to the group size hyperparameter, $G$, by evaluating two distinct values.

Finally, to solidify the reproducibility and statistical stability of our findings, all experiments are carried out across two independent random seeds. For each training run, we monitor four evaluation-time metrics (accuracy (eval), total reward (eval), entropy (eval), completion length (eval)) and four training-time metrics (reward (train), reward coefficient of variance (CV) (train), KL divergence (train), and completion clipped ratio (train)). For out-of-distribution evaluation, we have used different benchmarks: for GSM8K, we have used GSM-Plus \cite{li2024gsm}, Math-500 \cite{lightman2023letsverifystepstep}, AIME-2026 \cite{balunovic_srimatharena_2025}, MetaMathQA \cite{yu2023metamath}, NuminaMath \cite{numina_math_datasets}, OpenMath2 \cite{toshniwal2024openmath2}; for ARC, we have used MMLU-STEM \cite{mmlu}, BigBenchHard \cite{suzgun2022challenging}, LogiQA\cite{liu2020logiqa}, and GPQA-main\cite{rein2023gpqa}; and for OpenCoder, we have used MBPP \cite{austin2021program} and CodeParrot APPS \cite{hendrycksapps2021}, and Open-R1 Codeforces \cite{penedo2025codeforces}. The hyperparameters used are mentioned in Appendix \ref{app:params}. To handle reward sparsity, we have used reward functions mentioned in Appendix \ref{app:rewards}.

\section{Results}
\subsection{GRPO Training Dynamics on GSM8K}

\begin{figure}[h]
    \centering
    \includegraphics[width=\linewidth]{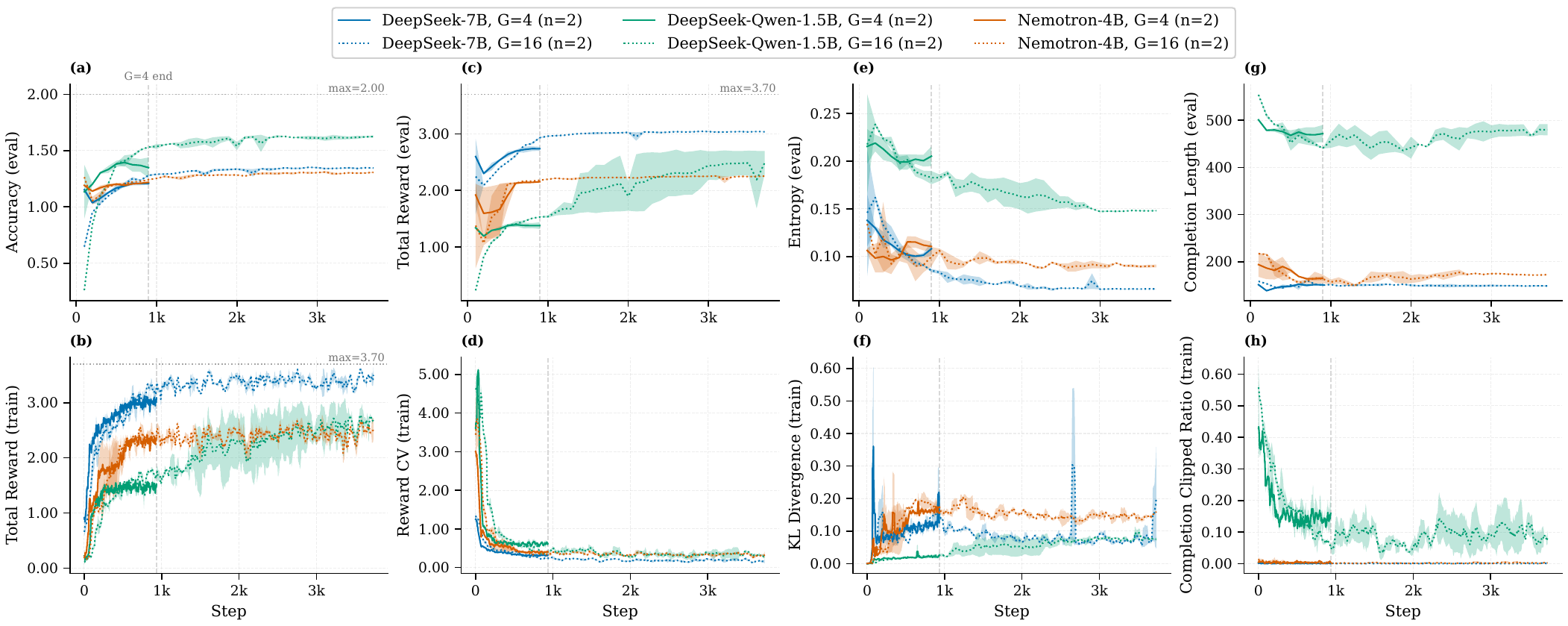}
    \caption{GRPO training dynamics on GSM8K across models and group sizes ($G \in \{4,16\}$). Evaluation accuracy improves rapidly before plateauing, with DeepSeek-R1-Qwen-1.5B achieving the highest final accuracy, particularly under $G=16$. Reward and accuracy are only partially aligned: DeepSeek-7B obtains the highest training and evaluation reward, whereas DeepSeek-R1-Qwen-1.5B attains the best accuracy with lower reward. The diagnostics reveal model-dependent behavior: DeepSeek-R1-Qwen-1.5B maintains higher entropy, longer completions, lower KL, and more persistent clipping than DeepSeek-7B and Nemotron-4B.}
    \label{fig:grpo_dynamics_gsm8k}
\end{figure}

% Requires:
% \usepackage{booktabs}
% \usepackage{multirow}
% \usepackage{xcolor}
% \definecolor{darkgreen}{RGB}{0,128,0}

\begin{table}[h]
\centering
\caption{Out-of-distribution evaluation of GRPO models trained on GSM8K dataset across math benchmarks using pass@k (\%). Results are reported as mean $\pm$ standard deviation over random seeds. For MetaMathQA and OpenMath2 datasets, we randomly sample 2000 examples for evaluation. For NuminaMath, we skip one proof problem that is not objectively verifiable.}
\label{tab:gsm8k_benchmark}
\scriptsize
\setlength{\tabcolsep}{3pt}
\begin{tabular}{llccccccccc}
\toprule
& &
\multicolumn{3}{c}{\textbf{DeepSeek-7B}}
& \multicolumn{3}{c}{\textbf{DeepSeek-R1-Qwen-1.5B}}
& \multicolumn{3}{c}{\textbf{Nemotron-4B}} \\
\cmidrule(lr){3-5}
\cmidrule(lr){6-8}
\cmidrule(lr){9-11}
\textbf{Benchmark} & \textbf{Metric}
& \textbf{Base} & \textbf{G=16} & \textbf{G=4}
& \textbf{Base} & \textbf{G=16} & \textbf{G=4}
& \textbf{Base} & \textbf{G=16} & \textbf{G=4} \\
\midrule

\multirow{2}{*}{GSM-Plus (2400)}
& pass@1
& 25.64 & {\color{ForestGreen}30.83{\scriptsize$\pm$0.71}} & {\color{ForestGreen}40.25{\scriptsize$\pm$0.08}}
& 33.48 & {\color{ForestGreen}36.52{\scriptsize$\pm$1.32}} & {\color{ForestGreen}34.80{\scriptsize$\pm$1.50}}
& 9.11 & {\color{darkgreen}\textbf{20.32{\scriptsize$\pm$5.55}}} & {\color{ForestGreen}19.46{\scriptsize$\pm$2.92}} \\
& pass@5
& 49.25 & {\color{ForestGreen}52.19{\scriptsize$\pm$0.91}} & {\color{ForestGreen}60.50{\scriptsize$\pm$0.41}}
& 55.54 & {\color{ForestGreen}57.86{\scriptsize$\pm$0.67}} & {\color{ForestGreen}56.44{\scriptsize$\pm$1.15}}
& 21.92 & {\color{ForestGreen}40.86{\scriptsize$\pm$8.52}} & {\color{darkgreen}\textbf{42.02{\scriptsize$\pm$3.86}}} \\

\midrule
\multirow{2}{*}{Math-500 (500)}
& pass@1
& 15.00 & {\color{ForestGreen}16.12{\scriptsize$\pm$0.11}} & {\color{ForestGreen}15.48{\scriptsize$\pm$1.07}}
& 70.80 & {\color{ForestGreen}75.74{\scriptsize$\pm$0.59}} & {\color{ForestGreen}73.18{\scriptsize$\pm$0.08}}
& 11.64 & {\color{darkgreen}\textbf{13.34{\scriptsize$\pm$0.31}}} & {\color{ForestGreen}12.62{\scriptsize$\pm$0.93}} \\
& pass@5
& 32.40 & {\color{ForestGreen}33.50{\scriptsize$\pm$0.14}} & {\color{ForestGreen}33.30{\scriptsize$\pm$3.54}}
& 85.20 & {\color{darkgreen}\textbf{89.60{\scriptsize$\pm$0.85}}} & {\color{ForestGreen}88.70{\scriptsize$\pm$1.84}}
& 31.40 & {\color{ForestGreen}31.50{\scriptsize$\pm$0.42}} & {\color{ForestGreen}31.60{\scriptsize$\pm$1.98}} \\

\midrule
\multirow{2}{*}{AIME-2026 (30)}
& pass@1
& 0.00 & {0.00{\scriptsize$\pm$0.00}} & {\color{ForestGreen}0.34{\scriptsize$\pm$0.47}}
& 1.33 & {\color{darkgreen}\textbf{6.17{\scriptsize$\pm$0.23}}} & {\color{ForestGreen}4.50{\scriptsize$\pm$0.24}}
& 0.00 & {0.00{\scriptsize$\pm$0.00}} & {0.00{\scriptsize$\pm$0.00}} \\
& pass@5
& 0.00 & {0.00{\scriptsize$\pm$0.00}} & {\color{ForestGreen}1.67{\scriptsize$\pm$2.35}}
& 5.19 & {\color{darkgreen}\textbf{11.78{\scriptsize$\pm$0.66}}} & {\color{ForestGreen}9.82{\scriptsize$\pm$0.14}}
& 0.00 & {0.00{\scriptsize$\pm$0.00}} & {0.00{\scriptsize$\pm$0.00}} \\

\midrule
\multirow{2}{*}{MetaMathQA (2000)}
& pass@1
& 66.93 & {\color{ForestGreen}73.07{\scriptsize$\pm$0.42}} & {\color{ForestGreen}70.30{\scriptsize$\pm$0.91}}
& 82.71 & {\color{ForestGreen}86.93{\scriptsize$\pm$1.10}} & {\color{ForestGreen}85.01{\scriptsize$\pm$0.79}}
& 46.54 & {\color{darkgreen}\textbf{52.55{\scriptsize$\pm$0.01}}} & {\color{ForestGreen}49.73{\scriptsize$\pm$0.38}} \\
& pass@5
& 90.65 & {\color{ForestGreen}91.78{\scriptsize$\pm$0.46}} & {\color{ForestGreen}90.88{\scriptsize$\pm$0.81}}
& 95.95 & {\color{ForestGreen}96.53{\scriptsize$\pm$0.04}} & {\color{ForestGreen}96.45{\scriptsize$\pm$0.21}}
& 74.05 & {\color{darkgreen}\textbf{75.43{\scriptsize$\pm$0.04}}} & {\color{ForestGreen}75.15{\scriptsize$\pm$0.64}} \\

\midrule
\multirow{2}{*}{NuminaMath (99)}
& pass@1
& 16.97 & {\color{red}15.96{\scriptsize$\pm$0.86}} & {\color{ForestGreen}17.57{\scriptsize$\pm$1.14}}
& 42.42 & {\color{ForestGreen}45.96{\scriptsize$\pm$0.14}} & {\color{ForestGreen}44.95{\scriptsize$\pm$0.71}}
& 14.55 & {\color{darkgreen}\textbf{17.37{\scriptsize$\pm$2.86}}} & {\color{red}14.55{\scriptsize$\pm$2.28}} \\
& pass@5
& 30.30 & {\color{red}28.79{\scriptsize$\pm$0.71}} & {\color{darkgreen}\textbf{35.35{\scriptsize$\pm$0.00}}}
& 61.62 & {\color{red}60.61{\scriptsize$\pm$1.43}} & {\color{red}60.11{\scriptsize$\pm$0.71}}
& 33.33 & {\color{red}32.32{\scriptsize$\pm$2.86}} & {\color{red}30.81{\scriptsize$\pm$7.85}} \\

\midrule
\multirow{2}{*}{OpenMath2 (2000)}
& pass@1
& 26.88 & {\color{ForestGreen}29.26{\scriptsize$\pm$0.68}} & {\color{ForestGreen}27.84{\scriptsize$\pm$0.29}}
& 54.42 & {\color{ForestGreen}62.15{\scriptsize$\pm$0.09}} & {\color{ForestGreen}59.05{\scriptsize$\pm$1.12}}
& 20.91 & {\color{darkgreen}\textbf{24.52{\scriptsize$\pm$0.19}}} & {\color{ForestGreen}22.51{\scriptsize$\pm$0.59}} \\
& pass@5
& 49.15 & {\color{ForestGreen}49.65{\scriptsize$\pm$0.85}} & {\color{red}48.35{\scriptsize$\pm$0.78}}
& 75.95 & {\color{ForestGreen}79.58{\scriptsize$\pm$0.81}} & {\color{ForestGreen}77.88{\scriptsize$\pm$0.04}}
& 42.00 & {\color{darkgreen}\textbf{44.35{\scriptsize$\pm$0.28}}} & {\color{ForestGreen}42.73{\scriptsize$\pm$0.39}} \\

\midrule
\multirow{2}{*}{IMO-Bench (400)}
& pass@1
& 2.00 & {\color{ForestGreen}2.23{\scriptsize$\pm$0.32}} & {\color{ForestGreen}2.20{\scriptsize$\pm$0.07}}
& 3.75 & {\color{darkgreen}\textbf{4.88{\scriptsize$\pm$0.81}}} & {\color{ForestGreen}4.67{\scriptsize$\pm$0.46}}
& 2.35 & {\color{red}2.17{\scriptsize$\pm$0.11}} & {\color{darkgreen}\textbf{2.60{\scriptsize$\pm$0.64}}} \\
& pass@5
& 7.50 & {\color{ForestGreen}7.88{\scriptsize$\pm$1.24}} & {\color{darkgreen}\textbf{8.25{\scriptsize$\pm$0.71}}}
& 10.75 & {\color{darkgreen}\textbf{14.38{\scriptsize$\pm$2.65}}} & {\color{ForestGreen}13.38{\scriptsize$\pm$1.24}}
& 9.00 & {9.00{\scriptsize$\pm$0.35}} & {\color{red}8.88{\scriptsize$\pm$1.24}} \\

\midrule
\multirow{2}{*}{HMMT-2025 (30)}
& pass@1
& 0.00 & {\color{ForestGreen}0.34{\scriptsize$\pm$0.47}} & {0.00{\scriptsize$\pm$0.00}}
& 6.00 & {\color{ForestGreen}6.33{\scriptsize$\pm$0.47}} & {\color{darkgreen}\textbf{7.33{\scriptsize$\pm$0.94}}}
& 0.00 & {0.00{\scriptsize$\pm$0.00}} & {0.00{\scriptsize$\pm$0.00}} \\
& pass@5
& 0.00 & {\color{darkgreen}\textbf{1.67{\scriptsize$\pm$2.35}}} & {0.00{\scriptsize$\pm$0.00}}
& 10.00 & {\color{darkgreen}\textbf{16.67{\scriptsize$\pm$0.00}}} & {\color{ForestGreen}15.00{\scriptsize$\pm$2.36}}
& 0.00 & {0.00{\scriptsize$\pm$0.00}} & {0.00{\scriptsize$\pm$0.00}} \\

\bottomrule
\end{tabular}
\end{table}
Figure~\ref{fig:grpo_dynamics_gsm8k} summarizes GRPO training dynamics on GSM8K across model families, group sizes $G$, and seeds. The evaluation accuracy in Figure~\ref{fig:grpo_dynamics_gsm8k}(a) and the training reward in Figure~\ref{fig:grpo_dynamics_gsm8k}(b) improve rapidly during early optimization and then exhibit diminishing returns, with most gains occurring within the first $\sim$ 500 steps. However, reward and accuracy are only partially aligned: DeepSeek-R1-Qwen-1.5B with $G=16$ achieves the highest evaluation accuracy, whereas DeepSeek-7B attains the highest training and evaluation reward. This discrepancy suggests that the optimized reward is not a fully reliable proxy for accuracy on GSM8K, and there is further scope to improve reward design.

The effect of group size $G$ is governed by model capacity and advantage estimation stability. For 7B and 4B models, larger $G$ provides diminishing returns due to early reward saturation in higher-capacity policies. In contrast, the 1.5B model benefits from $G=16$, which provides a stronger learning signal under its constrained update dynamics and higher entropy. 

The diagnostic metrics reveal distinct effective optimization regimes across models. The reward coefficient of variation in Figure~\ref{fig:grpo_dynamics_gsm8k}(d) decreases sharply during early training and stabilizes after $\sim$ 500 steps, indicating reduced variability in group-level rewards. However, the CV remains non-zero throughout training, suggesting that the group-relative reward signal remains non-degenerate. The entropy in Figure~\ref{fig:grpo_dynamics_gsm8k}(e) decreases across models, reflecting increasingly concentrated output distributions, but remains highest for DeepSeek-R1-Qwen-1.5B. The KL divergence in Figure~\ref{fig:grpo_dynamics_gsm8k}(f) remains bounded overall but is model-dependent: DeepSeek-7B and Nemotron-4B exhibit larger KL movement and occasional spikes, whereas DeepSeek-R1-Qwen-1.5B maintains relatively low measured KL despite achieving the highest accuracy.

Finally, Figure~\ref{fig:grpo_dynamics_gsm8k}(g) shows pronounced differences in completion length. DeepSeek-R1-Qwen-1.5B produces substantially longer completions, roughly 450-500 tokens, whereas DeepSeek-7B and Nemotron-4B remain closer to $\sim$ 150-180 tokens. This difference is mirrored by the completion clipped ratio in Figure~\ref{fig:grpo_dynamics_gsm8k}(h): DeepSeek-7B and Nemotron-4B remain near zero, while DeepSeek-R1-Qwen-1.5B exhibits persistent clipping, especially early in training. We hypothesize this disparity stems from the fact that DeepSeek-R1-Distill-Qwen-1.5B is already fine-tuned for reasoning-intensive tasks. Together, these diagnostics show that GSM8K training is not governed solely by reward improvement; models differ substantially in reward--accuracy alignment, output concentration, verbosity, policy movement, and the fraction of updates operating in the clipped regime.

\paragraph{Benchmarking.}

Table~\ref{tab:gsm8k_benchmark} reports out-of-distribution math evaluation for GRPO-trained models relative to their corresponding base models. Green entries denote improvements over the base model, while red entries denote regressions; bold green marks the model with the largest relative improvement for each benchmark metric. Overall, GRPO improves performance in 58 out of 72 model--benchmark--metric configurations ($\sim 80\%$), and every benchmark metric has at least one GRPO-trained model that outperforms its base counterpart. This suggests that GRPO training often transfers beyond the GSM8K training distribution. Qualitative inspection of generated solutions (Appendix\ref{appendix:gsm}) shows the improvement in answer quality for math reasoning problems. 

The gains are nevertheless heterogeneous. DeepSeek-R1-Qwen-1.5B achieves the strongest absolute performance on several benchmarks and is the only model family with substantial gains on AIME-2026. Nemotron-4B obtains the largest relative improvement in 8 out of 12 benchmark metrics, reflecting larger headroom from its lower base performance. Larger group size is not uniformly beneficial: $G=16$ often helps DeepSeek-R1-Qwen-1.5B, whereas several DeepSeek-7B and Nemotron-4B results are comparable or better under $G=4$. Thus, GRPO improves OOD math generalization in most cases, but its effectiveness depends on the base model, target benchmark, and group size.

\subsection{GRPO Training Dynamics on ARC}

\begin{figure}[H]
    \centering
    \includegraphics[width=\linewidth]{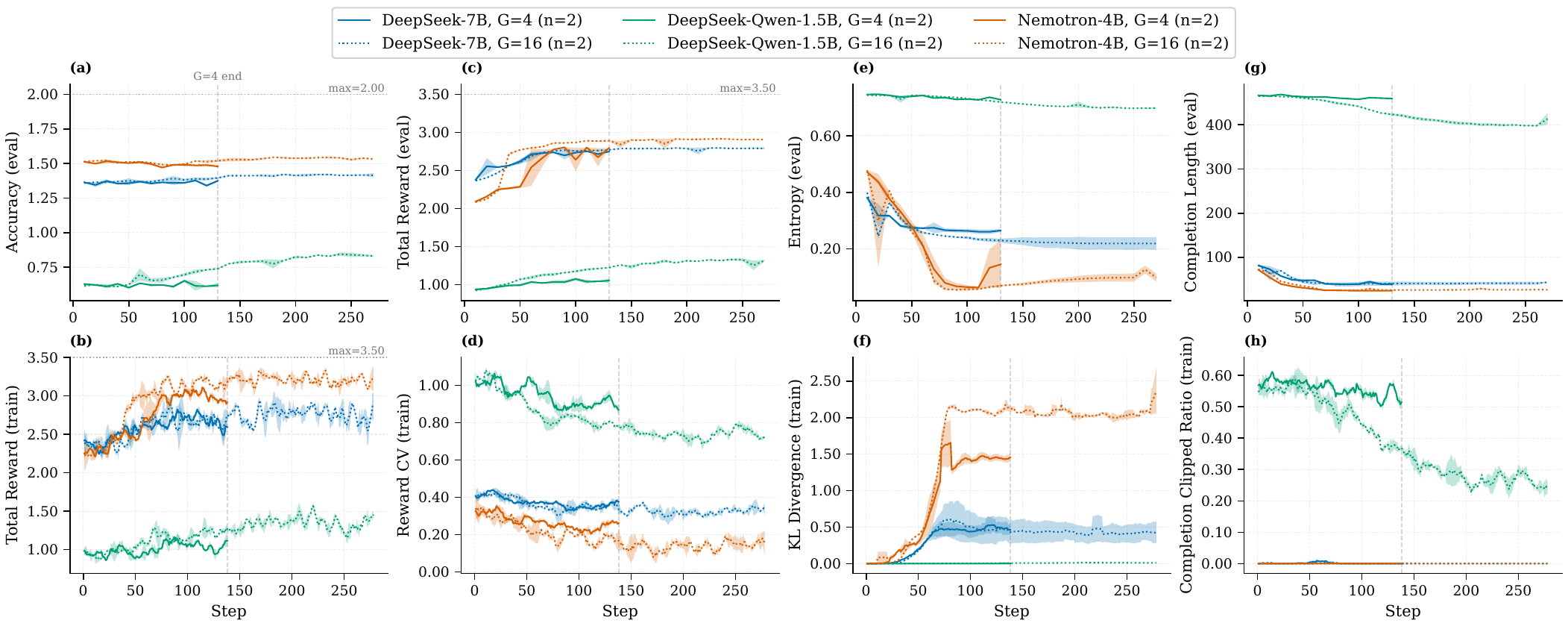}
    \caption{
    GRPO training dynamics on ARC across models and group sizes ($G \in \{4,16\}$). Reward and accuracy are largely aligned, with Nemotron-4B performing best, followed by DeepSeek-7B and DeepSeek-R1-Qwen-1.5B. 
Increasing $G$ yields limited gains for the stronger models but modestly improves DeepSeek-R1-Qwen-1.5B. Diagnostics reveal model-dependent regimes: Nemotron-4B exhibits low entropy with large KL movement, whereas DeepSeek-R1-Qwen-1.5B maintains high entropy, long completions, and persistent clipping.
    }
    \label{fig:grpo_dynamics_arc}
\end{figure}

Figure~\ref{fig:grpo_dynamics_arc} summarizes GRPO training dynamics on ARC. 
The evaluation accuracy in Figure~\ref{fig:grpo_dynamics_arc}(a) changes modestly over training, while the training reward in Figure~\ref{fig:grpo_dynamics_arc}(b) increases more substantially before plateauing. 
Nemotron-4B performs best, followed by DeepSeek-7B and DeepSeek-R1-Qwen-1.5B on both evaluation accuracy and reward. This suggests stronger reward--accuracy alignment on ARC, with DeepSeek-R1-Qwen-1.5B remaining separated from the other two models across both metrics.

Increasing the group size from $G=4$ to $G=16$ has a varied effect. 
For DeepSeek-7B and Nemotron-4B, a larger $G$ yields limited additional improvement in final accuracy or reward, indicating diminishing returns relative to the increased sampling cost. 
For DeepSeek-R1-Qwen-1.5B, however, $G=16$ improves both reward and accuracy over $G=4$, although the gap to the stronger models remains large. 
The consistency between training and evaluation rewards in Figures~\ref{fig:grpo_dynamics_arc}(b) and (c) further suggests that reward gains transfer to evaluation more reliably in this setting.

The diagnostic metrics reveal distinct effective optimization regimes across models. 
The reward coefficient of variation in Figure~\ref{fig:grpo_dynamics_arc}(d) decreases over training but remains non-zero, indicating that sampled completions continue to receive distinguishable rewards and that the group-relative advantage signal does not fully collapse. 
The entropy in Figure~\ref{fig:grpo_dynamics_arc}(e) decreases sharply for Nemotron-4B and more gradually for DeepSeek-7B, reflecting increasingly concentrated output distributions, while DeepSeek-R1-Qwen-1.5B maintains substantially higher entropy throughout training. 
The KL divergence in Figure~\ref{fig:grpo_dynamics_arc}(f) is also strongly model-dependent: it remains near zero for DeepSeek-R1-Qwen-1.5B, and increasing for DeepSeek-7B and Nemotron-4B, especially under $G=16$.

Finally, Figure~\ref{fig:grpo_dynamics_arc}(g) shows pronounced differences in response length, similar to GSM8K dataset. This is a inherent property of model. 
These diagnostics show that ARC training is not governed solely by reward improvement; rather, models differ substantially in policy movement, output concentration, verbosity, and the fraction of updates operating in the clipped regime.

\paragraph{Benchmarking.}
Table~\ref{tab:mmlu_bbh_logiqa_arc_gpqa_benchmark} reports out-of-distribution MCQ evaluation for GRPO-trained models relative to their corresponding base models. Overall, GRPO improves performance in 28 out of 72 model--benchmark--metric configurations ($\sim39\%$), and nearly every benchmark metric contains at least one GRPO-trained configuration that outperforms its corresponding base model. The gains, however, are highly heterogeneous across model families and evaluation settings. DeepSeek-R1-Qwen-1.5B achieves the strongest overall performance on several benchmarks and is the only model family exhibiting substantial improvements on GPQA-Main. It also attains the most frequent improvements (16 out of 24 benchmark metrics), potentially reflecting greater optimization headroom from its lower baseline performance. Larger group size is generally beneficial (e.g., $G=16$ improves GPQA-Main performance for Nemotron-4B). Overall, these results suggest that GRPO can improve out-of-distribution MCQ generalization, but its effectiveness remains strongly dependent on the underlying model family and training configuration.
\vspace{-1em}
% 

% Requires:
% \usepackage{booktabs}
% \usepackage{multirow}
% \usepackage[dvipsnames]{xcolor}   % dvipsnames needed for ForestGreen
% \definecolor{darkgreen}{RGB}{0,128,0}

\begin{table}[h]
\centering
\caption{Out-of-distribution evaluation of models trained on ARC dataset across Reasoning benchmarks using pass@k (\%). Results are reported as mean $\pm$ standard deviation over random seeds. For BigBenchHard (BBH), we use 4 subsets: logical\_deduction\_five\_objects, tracking\_shuffled\_objects\_five\_objects, formal\_fallacies, and temporal\_sequences to ensure a diversity of questions.}
\label{tab:mmlu_bbh_logiqa_arc_gpqa_benchmark}
\scriptsize
\setlength{\tabcolsep}{3pt}
\begin{tabular}{llccccccccc}
\toprule
& &
\multicolumn{3}{c}{\textbf{DeepSeek-7B}}
& \multicolumn{3}{c}{\textbf{DeepSeek-R1-Qwen-1.5B}}
& \multicolumn{3}{c}{\textbf{Nemotron-4B}} \\
\cmidrule(lr){3-5}
\cmidrule(lr){6-8}
\cmidrule(lr){9-11}
\textbf{Benchmark} & \textbf{Metric}
& \textbf{Base} & \textbf{G=16} & \textbf{G=4}
& \textbf{Base} & \textbf{G=16} & \textbf{G=4}
& \textbf{Base} & \textbf{G=16} & \textbf{G=4} \\
\midrule

\multirow{3}{*}{MMLU-STEM (3153)}
& pass@1
& 40.14 & {\color{darkgreen}\textbf{41.64{\scriptsize$\pm$0.25}}} & {\color{ForestGreen}40.97{\scriptsize$\pm$0.01}}
& 56.18 & {\color{red}55.44{\scriptsize$\pm$0.18}} & {\color{red}55.23{\scriptsize$\pm$0.05}}
& 48.56 & {\color{red}47.82{\scriptsize$\pm$0.05}} & {\color{red}46.92{\scriptsize$\pm$0.16}} \\

& pass@3
& 59.95 & {\color{red}57.09{\scriptsize$\pm$0.66}} & {\color{red}57.82{\scriptsize$\pm$0.36}}
& 79.31 & {\color{darkgreen}\textbf{79.82{\scriptsize$\pm$0.05}}} & {\color{ForestGreen}79.73{\scriptsize$\pm$0.03}}
& 62.92 & {\color{red}58.17{\scriptsize$\pm$0.09}} & {\color{red}59.64{\scriptsize$\pm$0.31}} \\

& pass@5
& 68.79 & {\color{red}63.62{\scriptsize$\pm$0.64}} & {\color{red}65.16{\scriptsize$\pm$0.59}}
& 86.36 & {\color{darkgreen}\textbf{87.33{\scriptsize$\pm$0.21}}} & {\color{ForestGreen}87.28{\scriptsize$\pm$0.16}}
& 68.92 & {\color{red}62.43{\scriptsize$\pm$0.11}} & {\color{red}65.23{\scriptsize$\pm$0.52}} \\

\midrule

\multirow{3}{*}{BBH (1000)}
& pass@1
& 30.82 & {\color{ForestGreen}30.83{\scriptsize$\pm$0.21}} & {\color{ForestGreen}31.12{\scriptsize$\pm$0.02}}
& 40.76 & {\color{darkgreen}\textbf{41.97{\scriptsize$\pm$0.55}}} & {\color{ForestGreen}41.72{\scriptsize$\pm$0.16}}
& 31.40 & {\color{red}30.67{\scriptsize$\pm$0.85}} & {\color{red}30.78{\scriptsize$\pm$0.54}} \\

& pass@3
& 51.17 & {\color{red}46.78{\scriptsize$\pm$1.24}} & {\color{red}47.95{\scriptsize$\pm$0.30}}
& 69.43 & {\color{red}69.04{\scriptsize$\pm$0.24}} & {\color{red}69.14{\scriptsize$\pm$0.12}}
& 51.45 & {\color{red}44.80{\scriptsize$\pm$1.91}} & {\color{red}49.58{\scriptsize$\pm$0.71}} \\

& pass@5
& 60.70 & {\color{red}53.35{\scriptsize$\pm$1.65}} & {\color{red}55.15{\scriptsize$\pm$0.25}}
& 80.20 & {\color{red}79.10{\scriptsize$\pm$0.20}} & {\color{red}78.70{\scriptsize$\pm$0.10}}
& 61.10 & {\color{red}51.65{\scriptsize$\pm$2.65}} & {\color{red}58.65{\scriptsize$\pm$0.85}} \\

\midrule

\multirow{3}{*}{LogiQA (651)}
& pass@1
& 37.70 & {\color{red}37.38{\scriptsize$\pm$0.02}} & {\color{red}37.66{\scriptsize$\pm$0.09}}
& 33.21 & {\color{red}32.09{\scriptsize$\pm$0.32}} & {\color{red}32.09{\scriptsize$\pm$0.54}}
& 35.94 & {\color{darkgreen}\textbf{37.25{\scriptsize$\pm$0.35}}} & {\color{ForestGreen}36.17{\scriptsize$\pm$0.32}} \\

& pass@3
& 51.18 & {\color{red}47.54{\scriptsize$\pm$0.26}} & {\color{red}49.88{\scriptsize$\pm$0.00}}
& 60.18 & {\color{darkgreen}\textbf{60.65{\scriptsize$\pm$0.07}}} & {\color{ForestGreen}60.59{\scriptsize$\pm$0.44}}
& 52.67 & {\color{red}49.80{\scriptsize$\pm$0.91}} & {\color{ForestGreen}52.87{\scriptsize$\pm$0.63}} \\

& pass@5
& 57.14 & {\color{red}52.15{\scriptsize$\pm$0.38}} & {\color{red}54.84{\scriptsize$\pm$0.15}}
& 72.66 & {\color{ForestGreen}73.19{\scriptsize$\pm$0.38}} & {\color{darkgreen}\textbf{73.50{\scriptsize$\pm$0.38}}}
& 60.22 & {\color{red}55.14{\scriptsize$\pm$0.93}} & {\color{red}60.06{\scriptsize$\pm$1.07}} \\

\midrule

\multirow{3}{*}{GPQA-Main (448)}
& pass@1
& 20.58 & {\color{ForestGreen}21.56{\scriptsize$\pm$0.40}} & {\color{ForestGreen}21.77{\scriptsize$\pm$0.34}}
& 7.90 & {\color{darkgreen}\textbf{10.69{\scriptsize$\pm$0.02}}} & {\color{ForestGreen}9.05{\scriptsize$\pm$0.33}}
& 14.20 & {\color{ForestGreen}18.46{\scriptsize$\pm$0.25}} & {\color{red}12.43{\scriptsize$\pm$0.16}} \\

& pass@3
& 41.27 & {\color{red}39.39{\scriptsize$\pm$0.01}} & {\color{ForestGreen}42.45{\scriptsize$\pm$0.14}}
& 15.22 & {\color{darkgreen}\textbf{18.34{\scriptsize$\pm$0.19}}} & {\color{ForestGreen}16.61{\scriptsize$\pm$0.38}}
& 30.89 & {\color{ForestGreen}35.83{\scriptsize$\pm$0.93}} & {\color{red}27.66{\scriptsize$\pm$0.53}} \\

& pass@5
& 52.90 & {\color{red}48.11{\scriptsize$\pm$1.00}} & 52.90{\scriptsize$\pm$0.22}
& 19.64 & {\color{ForestGreen}21.88{\scriptsize$\pm$0.00}} & {\color{ForestGreen}20.88{\scriptsize$\pm$0.34}}
& 41.07 & {\color{darkgreen}\textbf{46.32{\scriptsize$\pm$1.45}}} & {\color{red}37.61{\scriptsize$\pm$0.78}} \\

\bottomrule
\end{tabular}
\end{table}

\vspace{-1em}
\paragraph{Refinement Treatments.} We use mechanistic evaluation metrics to decide on the choice of refinement treatment for Nemotron-4B and DeepSeek-7B. Table \ref{tab:combined_hyperparameter_refinement} shows different refinement treatments employed on Nemotron-4B: Reduced Layer Count (R. Layer), Reduced Module (R. Module), $\beta=0.01$ (0.003 is typical baseline value), and a combination of all three. We choose to increase $\beta$ after observing high KL-divergence (train) for Nemotron-4B. In R. Layer, we choose to fine-tune only the first and last two layers (four in total out of 32 layers). This is informed by the weight displacement and activation shift analysis (Figures \ref{fig:weight_delta_heatmap} and \ref{fig:logit_shift}). From Table \ref{tab:combined_hyperparameter_refinement}, we observe improvements in performance with lower performance degradation from the base model. 

Table \ref{tab:combined_hyperparameter_refinement} shows the treatment of DeepSeek-7B by changing clipping coefficient from 0.2 to 0.3, in order to improve the sample diversity (Figure \ref{fig:grpo_dynamics_arc} shows relatively high reward but low coefficient of variance). We observe in many cases it is getting nearer to the base model. Empirical examples (Appendix \ref{appendix:arc}) corroborate these findings.

\begin{table}[h]
\centering
\caption{Different hyperparameter refinement treatments for improving fine-tuning
performance (pass@k score) on MCQ benchmarks. DeepSeek-7B treatment changes the
clipping coefficient from $\varepsilon_{high}=0.2$ to $0.3$. Nemotron-4B treatments
include reducing LoRA fine-tuning layers (R. Layer), reducing LoRA fine-tuning modules
(R. Module), increasing KL regularization from $\beta=0.003$ to $0.01$, and combining all refinements.
GRPO-baseline ($G=4$) reports the mean over random seeds. Overall, we observe hyperparameter refinements improving GRPO-tuned model performance.}
\label{tab:combined_hyperparameter_refinement}

\scriptsize
\setlength{\tabcolsep}{3pt}

\resizebox{\textwidth}{!}{%
\begin{tabular}{llccccccccc}
\toprule
& &
\multicolumn{3}{c}{\textbf{DeepSeek-7B}}
& \multicolumn{6}{c}{\textbf{Nemotron-4B}} \\
\cmidrule(lr){3-5}
\cmidrule(lr){6-11}

\textbf{Benchmark} & \textbf{Metric}
& \textbf{Base}
& \textbf{G=4}
& \makecell[b]{\textbf{$\varepsilon_{high}=0.3$}\\\textbf{(G=4)}}
& \textbf{Base}
& \textbf{G=4}
& \makecell[b]{\textbf{R. Layer}\\\textbf{(G=4)}}
& \makecell[b]{\textbf{R. Module}\\\textbf{(G=4)}}
& \makecell[b]{\textbf{$\beta=0.01$}\\\textbf{(G=4)}}
& \makecell[b]{\textbf{All}\\\textbf{(G=4)}} \\
\midrule

\multirow{3}{*}{MMLU-STEM (3153)}
& pass@1
& 40.14
& {\color{ForestGreen}40.97 }
& {\color{ForestGreen}40.29 }
& 48.56
& {\color{red}46.92 }
& {\color{red}47.76 }
& {\color{red}48.21 }
& {\color{red}47.57 }
& {\color{red}48.37 } \\

& pass@3
& 59.95
& {\color{red}57.82 }
& {\color{red}59.65 }
& 62.92
& {\color{red}59.64 }
& {\color{red}62.27 }
& {\color{red}62.12 }
& {\color{red}61.12 }
& {\color{red}62.56 } \\

& pass@5
& 68.79
& {\color{red}65.16 }
& {\color{red}68.22 }
& 68.92
& {\color{red}65.23 }
& {\color{red}68.25 }
& {\color{red}67.90 }
& {\color{red}66.73 }
& {\color{red}68.60 } \\

\midrule

\multirow{3}{*}{BBH (1000)}
& pass@1
& 30.82
& {\color{ForestGreen}31.12 }
& {\color{ForestGreen}30.98 }
& 31.40
& {\color{red}30.78 }
& {\color{red}30.66 }
& {\color{red}30.80 }
& {\color{red}31.04 }
& {\color{red}30.68 } \\

& pass@3
& 51.17
& {\color{red}47.95 }
& {\color{red}50.69 }
& 51.45
& {\color{red}49.58 }
& {\color{ForestGreen}51.54 }
& {\color{red}50.90 }
& {\color{red}50.78 }
& {\color{red}49.55 } \\

& pass@5
& 60.70
& {\color{red}55.15 }
& {\color{red}59.70 }
& 61.10
& {\color{red}58.65 }
& {\color{ForestGreen}62.20 }
& {\color{ForestGreen}61.30 }
& {\color{red}60.20 }
& {\color{red}58.70 } \\

\midrule

\multirow{3}{*}{LogiQA (651)}
& pass@1
& 37.70
& {\color{red}37.67 }
& {\color{red}37.45 }
& 35.94
& {\color{ForestGreen}36.17 }
& {\color{ForestGreen}36.13 }
& {\color{ForestGreen}36.41 }
& {\color{red}35.82 }
& {\color{ForestGreen}35.98 } \\

& pass@3
& 51.18
& {\color{red}49.88 }
& {\color{red}50.63 }
& 52.67
& {\color{ForestGreen}52.87 }
& {\color{ForestGreen}53.06 }
& {\color{red}52.52 }
& {\color{ForestGreen}52.76 }
& {\color{ForestGreen}52.84 } \\

& pass@5
& 57.14
& {\color{red}54.84 }
& {\color{red}56.68 }
& 60.22
& {\color{red}60.06 }
& {\color{red}59.91 }
& {\color{red}58.53 }
& {\color{red}59.60 }
& {\color{ForestGreen}60.52 } \\

\midrule

\multirow{3}{*}{GPQA-Main (448)}
& pass@1
& 20.58
& {\color{ForestGreen}21.77 }
& {\color{ForestGreen}21.25 }
& 14.20
& {\color{red}12.43 }
& {\color{red}13.53 }
& {\color{red}13.17 }
& {\color{red}13.79 }
& {\color{red}13.44 } \\

& pass@3
& 41.27
& {\color{ForestGreen}42.46 }
& {\color{red}40.45 }
& 30.89
& {\color{red}27.66 }
& {\color{red}30.36 }
& {\color{red}29.40 }
& {\color{ForestGreen}30.92 }
& {\color{red}30.07 } \\

& pass@5
& 52.90
& 52.90 
& {\color{red}49.55 }
& 41.07
& {\color{red}37.61 }
& 41.07 
& {\color{red}40.40 }
& {\color{ForestGreen}41.52 }
& {\color{red}40.85 } \\

\bottomrule
\end{tabular}%
}
\end{table}

\subsection{GRPO Training Dynamics for OpenCoder}

\begin{figure}[b]
    \centering
    \includegraphics[width=\linewidth]{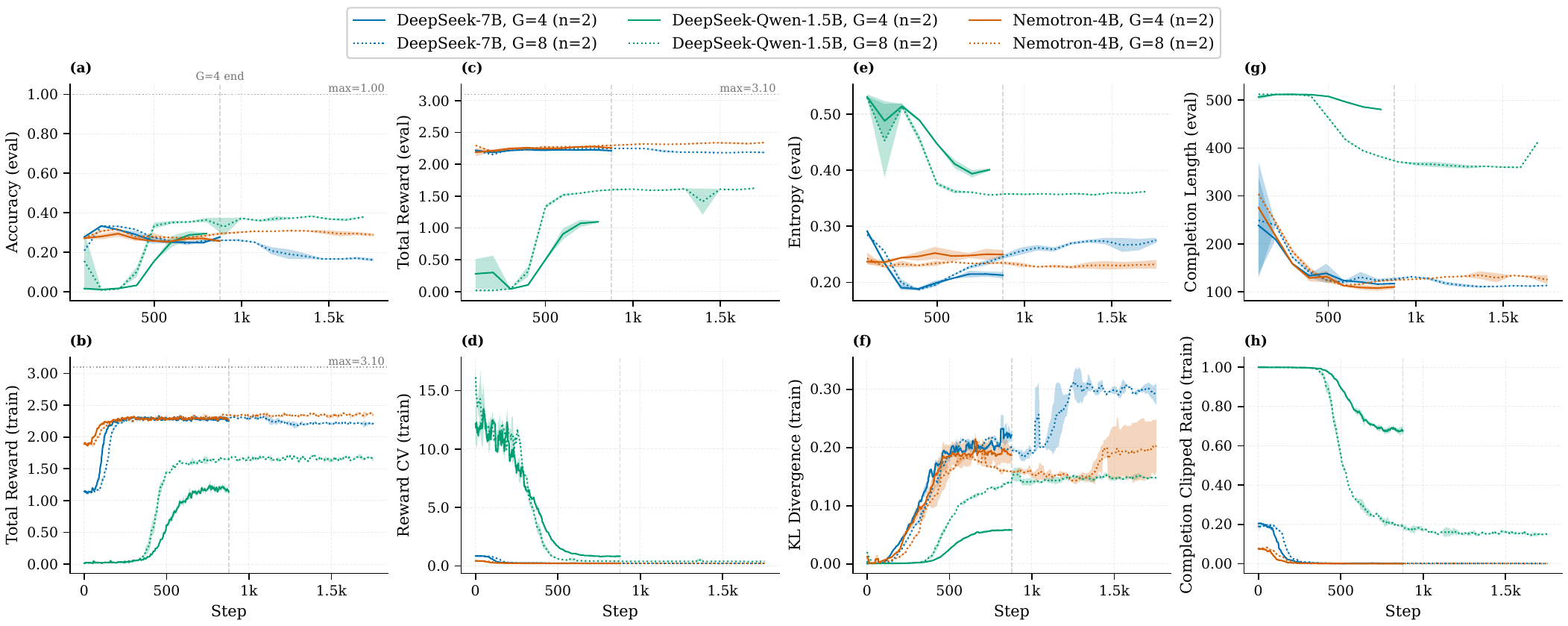}
    \caption{
     GRPO training dynamics on OpenCoder across models and group sizes ($G \in \{4,8\}$). 
Performance and reward improve rapidly before reaching diminishing returns, while larger $G$ provides only model-dependent gains. 
KL remains bounded, entropy generally decreases, and completion lengths vary substantially across models. 
}
    \label{fig:grpo_dynamics_OpenCoder}
\end{figure}

Figure~\ref{fig:grpo_dynamics_OpenCoder} summarizes GRPO training dynamics on OpenCoder across model families, group sizes $G$, and seeds. The evaluation accuracy in Figure~\ref{fig:grpo_dynamics_OpenCoder}(a) and the training reward in Figure~\ref{fig:grpo_dynamics_OpenCoder}(b) improve rapidly during early optimization and then exhibit diminishing returns, with most gains occurring within the first $\sim$ 500-1000 steps. 
% However, the reward and the accuracy metrics are not perfectly aligned: DeepSeek-R1-Qwen-1.5B with $G=8$ achieves the highest evaluation accuracy, whereas Nemotron-4B and DeepSeek-7B obtain substantially higher reward values. This discrepancy suggests that the GRPO reward is only a partial proxy for downstream accuracy.

% For DeepSeek-7B and Nemotron-4B, larger $G$ produces limited additional training-reward gains, as both models approach saturation early. For DeepSeek-R1-Qwen-1.5B, however, the $G=8$ run continues improving after $\sim$400 steps and reaches a higher final training reward than $G=4$. A similar separation is visible in evaluation accuracy, where DeepSeek-R1-Qwen-1.5B benefits most from the larger group size. By contrast, DeepSeek-7B shows a later decline in accuracy, coinciding with larger KL divergence in Figure~\ref{fig:grpo_dynamics_OpenCoder}(f).
The performance drop in the 7B model, accompanied by increased KL divergence in Figure~\ref{fig:grpo_dynamics_OpenCoder}, indicates that excessive optimization pressure can induce mode collapse, where the policy overfits to superficial high-reward patterns at the expense of pretrained reasoning. The diagnostic metrics reveal substantial differences in effective optimization behavior. 
The reward coefficient of variation in Figure~\ref{fig:grpo_dynamics_OpenCoder}(d) decreases sharply for DeepSeek-R1-Qwen-1.5B and remains low for DeepSeek-7B and Nemotron-4B, suggesting that sampled completions receive increasingly similar rewards and that the group-relative advantage signal becomes less variable. The entropy in Figure~\ref{fig:grpo_dynamics_OpenCoder}(e) generally decreases over training, indicating reduced sampling diversity, while KL divergence remains bounded, suggesting no clear policy instability under these settings. Finally, Figure~\ref{fig:grpo_dynamics_OpenCoder}(g) shows strong model-dependent differences in completion length.

\paragraph{Benchmarking.} Table~\ref{tab:code_benchmark} reports out-of-distribution evaluation on three code-generation benchmarks: MBPP, CodeParrot APPS (introductory subset), and Open-R1 Codeforces (rating 800 - 1600). Transfer to code generation is substantially more heterogeneous than to mathematical reasoning. GRPO improves performance in 17 of 36 model--benchmark--metric configurations, and the gains are primarily driven by two of the three models. First, DeepSeek-R1-Qwen-1.5B exhibits consistent and significant improvements across all three benchmarks, with pass@1 gains of +4.86\% on MBPP. Second, DeepSeek-7B achieves strongest transfer on the most challenging benchmark, Codeforces, where $G=8$ improves pass@1 from 0.76 to 2.60. In contrast, Nemotron-4B generally regresses on MBPP and APPS. 

Despite achieving the largest quantitative gains, DeepSeek-R1-Qwen-1.5B exhibits notable structural inconsistencies in its outputs. The model rarely produced well-formed \texttt{<reasoning>} tags and instead defaults to \texttt{<think>} tags with verbose, often repetitive chain-of-thought traces that overflow the token budget. Its generated answers are seldom wrapped in explicit \texttt{<answer>} tags, and code solutions tend to be embedded within large unstructured blocks. However, the model reliably produced proper code fences, which facilitates straightforward extraction of the final solution.

In contrast, Nemotron-4B and DeepSeek-7B produce more structurally consistent outputs and adhere more closely to the expected formatting schema. Qualitative inspection of generated solutions (see Appendices~\ref{apps:mbpp_examples} and~\ref{apps:apps_example}) reveals that the GRPO-trained variants of these models exhibit stylistic shifts towards more efficient code and targeted use of built-in library functions, patterns largely absent in their base counterparts. This suggests that GRPO can also shape solution quality along with correctness, even when aggregate pass@k improvements are modest.
\vspace{-1em}
\begin{table}[h]
\centering
\caption{Out-of-distribution evaluation of models trained on OpenCoder dataset across coding benchmarks using pass@k (\%). Results are reported as mean $\pm$ standard deviation over random seeds}
\label{tab:code_benchmark}
\scriptsize
\setlength{\tabcolsep}{3pt}
\begin{tabular}{llccccccccc}
\toprule
& &
\multicolumn{3}{c}{\textbf{DeepSeek-7B}}
& \multicolumn{3}{c}{\textbf{DeepSeek-R1-Qwen-1.5B}}
& \multicolumn{3}{c}{\textbf{Nemotron-4B}} \\
\cmidrule(lr){3-5}
\cmidrule(lr){6-8}
\cmidrule(lr){9-11}
\textbf{Benchmark} & \textbf{Metric}
& \textbf{Base} & \textbf{G=8} & \textbf{G=4}
& \textbf{Base} & \textbf{G=8} & \textbf{G=4}
& \textbf{Base} & \textbf{G=8} & \textbf{G=4} \\
\midrule

\multirow{2}{*}{MBPP (500)}
& pass@1
& 34.98
& {\color{red}34.87{\scriptsize$\pm$0.78}}
& {\color{ForestGreen}35.73{\scriptsize$\pm$0.21}}
& 32.38
& {\color{ForestGreen}34.46{\scriptsize$\pm$0.34}}
& {\color{darkgreen}\textbf{37.24{\scriptsize$\pm$0.14}}}
& 32.70
& {\color{ForestGreen}33.83{\scriptsize$\pm$0.64}}
& {\color{red}32.23{\scriptsize$\pm$0.30}} \\

& pass@5
& 51.27
& {\color{red}47.01{\scriptsize$\pm$0.72}}
& {\color{red}49.40{\scriptsize$\pm$1.09}}
& 49.51
& {\color{ForestGreen}53.48{\scriptsize$\pm$0.08}}
& {\color{darkgreen}\textbf{54.43{\scriptsize$\pm$0.04}}}
& 48.30
& {\color{red}42.17{\scriptsize$\pm$0.57}}
& {\color{red}41.29{\scriptsize$\pm$0.48}} \\

% & pass@10
% & 57.60
% & {\color{red}50.40{\scriptsize$\pm$0.28}}
% & {\color{red}54.40{\scriptsize$\pm$1.13}}
% & 55.00
% & {\color{ForestGreen}58.20{\scriptsize$\pm$0.00}}
% & {\color{darkgreen}\textbf{59.10{\scriptsize$\pm$0.71}}}
% & 54.00
% & {\color{red}45.00{\scriptsize$\pm$0.28}}
% & {\color{red}44.50{\scriptsize$\pm$0.14}} \\

\midrule

\multirow{2}{*}{APPS (1000)}
& pass@1
& 10.23
& {\color{red}9.33{\scriptsize$\pm$0.17}}
& {\color{ForestGreen}10.58{\scriptsize$\pm$0.37}}
& 20.90
& {\color{darkgreen}\textbf{26.14{\scriptsize$\pm$0.35}}}
& {\color{ForestGreen}24.14{\scriptsize$\pm$0.16}}
& 8.79
& {\color{red}6.59{\scriptsize$\pm$0.69}}
& {\color{red}6.28{\scriptsize$\pm$0.39}} \\

& pass@5
& 18.35
& {\color{red}15.37{\scriptsize$\pm$0.13}}
& {\color{red}16.85{\scriptsize$\pm$0.46}}
& 34.27
& {\color{darkgreen}\textbf{40.39{\scriptsize$\pm$0.33}}}
& {\color{ForestGreen}37.48{\scriptsize$\pm$0.56}}
& 19.86
& {\color{red}12.56{\scriptsize$\pm$1.65}}
& {\color{red}13.28{\scriptsize$\pm$1.13}} \\

% & pass@10
% & 22.80
% & {\color{red}18.20{\scriptsize$\pm$0.28}}
% & {\color{red}19.60{\scriptsize$\pm$0.57}}
% & 38.90
% & {\color{darkgreen}\textbf{45.30{\scriptsize$\pm$0.14}}}
% & {\color{ForestGreen}41.80{\scriptsize$\pm$0.42}}
% & 24.50
% & {\color{red}15.15{\scriptsize$\pm$1.77}}
% & {\color{red}16.65{\scriptsize$\pm$1.34}} \\

\midrule

\multirow{2}{*}{Codeforces (198)}
& pass@1
& 0.76
& {\color{darkgreen}\textbf{2.60{\scriptsize$\pm$0.10}}}
& {\color{ForestGreen}1.70{\scriptsize$\pm$0.30}}
& 4.29
& {\color{ForestGreen}6.20{\scriptsize$\pm$0.10}}
& {\color{ForestGreen}5.10{\scriptsize$\pm$0.00}}
& 3.33
& {\color{red}2.20{\scriptsize$\pm$0.60}}
& {\color{red}2.30{\scriptsize$\pm$0.57}} \\

& pass@5
& 2.64
& {\color{darkgreen}\textbf{6.10{\scriptsize$\pm$0.50}}}
& {\color{ForestGreen}4.30{\scriptsize$\pm$0.30}}
& 5.67
& {\color{ForestGreen}10.90{\scriptsize$\pm$0.10}}
& {\color{ForestGreen}8.00{\scriptsize$\pm$0.10}}
& 6.33
& {\color{red}4.40{\scriptsize$\pm$0.00}}
& {\color{red}5.10{\scriptsize$\pm$0.68}} \\

% & pass@10
% & 4.04
% & {\color{ForestGreen}7.60{\scriptsize$\pm$0.70}}
% & {\color{ForestGreen}5.80{\scriptsize$\pm$0.40}}
% & 6.06
% & {\color{darkgreen}\textbf{13.10{\scriptsize$\pm$0.00}}}
% & {\color{ForestGreen}9.60{\scriptsize$\pm$0.00}}
% & 8.08
% & {\color{red}5.60{\scriptsize$\pm$0.00}}
% & {\color{red}6.80{\scriptsize$\pm$0.38}} \\

\bottomrule
\end{tabular}
\end{table}

\paragraph{Refinement Treatments.} Table~\ref{tab:nemotron_code_compact} report the effect of mechanistically-informed refinement on the out-of-distribution code evaluation for Nemotron-4B and DeepSeek-7B, respectively. Treatment selection is guided by analyzing diagnostic metrics (Figures \ref{fig:weight_delta_heatmap} and \ref{fig:logit_shift}). 

\begin{table}[h]
\centering
% \caption{Nemotron-4B out-of-distribution code evaluation}
\caption{Different hyperparameter refinement treatments for improving fine-tuning performance (pass@k score) on coding benchmarks. DeepSeek-7B treatment reduces LoRA fine-tuning layers (R. Layer). Nemotron-4B treatments include reducing LoRA fine-tuning layers (R. Layer), increasing weight of correctness reward on top it (R. Layer + RR), and reducing LoRA fine-tuning modules (R. Module). GRPO-baseline (G=4) reports the mean over random seeds. Overall, we observe hyperparameter refinements improving GRPO-tuned model performance.}
\label{tab:nemotron_code_compact}
\footnotesize\textbf{}
\setlength{\tabcolsep}{3pt}
\resizebox{\textwidth}{!}{%
\begin{tabular}{llccccccccc}
\toprule
& & \multicolumn{3}{c}{\textbf{DeepSeek-7B}} &
\multicolumn{5}{c}{\textbf{Nemotron-4B}}\\
\cmidrule(l{6pt}r{6pt}){3-5}
\cmidrule(l{6pt}r{6pt}){6-10}
\textbf{Benchmark}
& \textbf{Metric}
& \textbf{Base}
& \makecell[b]{\textbf{G=4}}
& \makecell[b]{\textbf{R. Layer}\\\textbf{(G=4)}}
& \textbf{Base}
& \makecell[b]{\textbf{G=4}}
& \makecell[b]{\textbf{R. Layer}\\\textbf{(G=4)}}
& \makecell[b]{\textbf{R. Layer + RR}\\\textbf{(G=4)}}
& \makecell[b]{\textbf{R. Module + RR}\\\textbf{(G=4)}} \\
\midrule
\multirow{2}{*}{MBPP (500)}
& pass@1 & 34.98
& {\color{ForestGreen}35.73 }
& {\color{ForestGreen}36.46 } & 32.70 & {\color{red}32.23 } & {\color{ForestGreen}32.86 } & {\color{ForestGreen}33.56 } & {\color{ForestGreen}33.92 } \\
& pass@5  & 51.27
& {\color{red}49.40 }
& {\color{ForestGreen}51.44 } & 48.30 & {\color{red}41.29 } & {\color{red}46.47 } & {\color{red}46.85 } & {\color{red}45.99 } \\
% & pass@10 & 54.00 & {\color{red}44.50 } & {\color{red}51.60 } & {\color{red}51.60 } & {\color{red}49.80 } \\
\midrule
\multirow{2}{*}{APPS (1000)}
& pass@1  & 10.23
& {\color{ForestGreen}10.58 }
& {\color{ForestGreen}10.80 } & 8.79  & {\color{red}6.28 } & {\color{ForestGreen}10.13 } & {\color{ForestGreen}10.35 } & {\color{ForestGreen}11.83 } \\
& pass@5  & 18.35
& {\color{red}16.85 }
& {\color{red}17.62 } & 19.86 & {\color{red}13.28 } & {\color{ForestGreen}20.65 } & {\color{ForestGreen}20.34 } & {\color{ForestGreen}20.31 } \\
% & pass@10 & 24.50 & {\color{red}16.65 } & {\color{ForestGreen}25.40 } & {\color{red}24.40 } & {\color{red}24.00 } \\
\midrule
\multirow{2}{*}{Codeforces (198)}
& pass@1  & 0.76
& {\color{ForestGreen}1.70 }
& {\color{black}0.76 } & 3.33 & {\color{red}2.30 } & {\color{red}2.83 } & {\color{red}3.18 } & {\color{red}3.03 } \\
& pass@5  & 2.64
& {\color{ForestGreen}4.30 }
& {\color{ForestGreen}2.86 } & 6.33 & {\color{red}5.10 } & {\color{red}5.16 } & {\color{red}4.50 } & {\color{red}5.26 } \\
% & pass@10 & 8.08 & {\color{red}6.80 } & {\color{red}6.06 } & {\color{red}5.05 } & {\color{red}6.13 } \\
\bottomrule
\end{tabular}}
\end{table}

The GRPO baseline for Nemotron-4B $G=4$ degrades performance relative to the base model on nearly all benchmark--metric pairs, showing severe regressions on APPS (pass@1: $-28.6\%$) and Codeforces (pass@1: $-30.9\%$). We apply three cumulative treatments informed by mechanistic diagnostics. First, guided by weight displacement, Reduced Layer (R. Layer) restricts LoRA adaptation to the first and last two transformer layers (4 of 28). This treatment alone recovers most of the regression on MBPP and APPS. Second, we reshape the rewards(R. Layer + RR), specifically update weight for code correctness from 1.0 to 2.0. This yields further gains, bringing APPS pass@1 to $+17.8\%$ and narrowing the Codeforces pass@1 gap to $-4.5\%$. Third, Reduced Module (R. Module + RR) restricts adaptation to the value and output projection matrices and achieves the strongest pass@1 on both MBPP (33.92, $+3.7\%$) and APPS (11.83, $+34.6\%$). However, the pass@5 metrics do not see similar gains, suggesting that the treatments improve greedy accuracy at some cost to sample diversity.

Similarly, the GRPO baseline for DeepSeek-7B already showed positive transfer on several metrics, particularly on Codeforces (pass@1: $+123.7\%$). The R. Layer treatment further improves pass@1 on MBPP and APPS. This pattern suggests that restricting trainable layers can stabilize transfer to benchmarks distributionally closer to the training data but can reduce gains on more challenging, distributionally distant tasks (e.g. Codeforces). Across both models, our mechanistically-guided refinement treatments convert regressions to improvements in many cases. The results validate the utility of weight displacement and activation shift diagnostics as actionable signals for hyperparameter and architecture decisions in GRPO post-training.

\section{Related Work}
 Prior work on RLHF has extensively analyzed optimization stability, reward modeling, and policy training dynamics, primarily within the context of PPO-based alignment pipelines \citep{ouyang2022training, zheng2023secretsrlhflargelanguage_ppo, wang2024secretsrlhflargelanguage_reward, peng2023stabilizing, wolf2025reward}. These studies underscore the critical role that a parameterized value function plays in stabilizing training and improving overall performance. Ensuring the stability and performance of GRPO remains an active area of research. While recent literature has introduced several new variants of the GRPO algorithm \cite{yu2025dapo, zheng2025groupsequencepolicyoptimization, liu2025understandingr1zeroliketrainingcritical}, optimizing their implementation remains challenging.  

To address the inherent instability of actor-only methods, recent work has begun to explore mechanistic evaluation as a complementary signal for hyperparameter optimization and training diagnostics in large-scale post-training pipelines \cite{rai2024practical, naseem2026mechanistic, salt2025hyperparameter}. This direction is exceptionally relevant for GRPO, where optimization stability and downstream performance are highly sensitive to a high-dimensional hyperparameter space and lack the grounding of a traditional baseline estimator \cite{deng2025effect}.

\section{Conclusion}

Through systematic characterization of GRPO for SLMs (1.5B - 7B) under single-node compute constraints, we find that GRPO effectively enhances mathematical reasoning, while MCQ and coding tasks require targeted hyperparameter refinement. Code reasoning responds well to these refinements, but MCQ performance shows limited improvement, possible due to the knowledge intensive nature of these tasks. We summarize our findings as follows:

% its application to coding and MCQ tasks is suboptimal and requires precise optimization. The core contribution of our empirical study lies in the transition from "black-box" hyperparameter tuning to mechanistic-guided optimization. By analyzing tensor-level update dynamics and weight displacement patterns, we uncovered several critical insights for the community:

\begin{itemize}[leftmargin=*]
    \item \textbf{Targeted Hyperparameter Optimization:} We established that layer-wise and module-level contributions in LoRA-based GRPO are highly non-uniform. Strategic selection of target modules, informed by weight displacement analysis, improves optimization efficiency and downstream generalization compared to uniform adaptation.
    
    % \item \textbf{The Reward-Shaping Trade-off:} We identified that while reward shaping is essential to overcome sparse signals in complex domains like code generation, it introduces a sensitivity threshold. Excessively large reward scales, while initially accelerating optimization, ultimately jeopardize training stability and convergence.

    % \item \textbf{Controlled Experimentation for GRPO Reward Shaping:} Guided by our update analysis, we successfully refined LoRA configurations and reward structures to reverse initial performance stagnation in MCQ and coding benchmarks, proving that mechanistic feedback is a viable signal for escaping local minima in RFT.

    \item \textbf{GRPO Training Guidance:} We propose a set of GRPO training signals including four evaluation-time outcome metrics (accuracy, eval. reward, entropy, completion length) and training-time process metrics (training reward, reward CV, KL divergence, completion clipped ratio). They help in diagnosing training bottlenecks such as reward/accuracy misalignment and taking corrective actions. For example, when reward CV is low, hyperparameters such as temperature can be increased to improve sample diversity.   
    
    % We propose reward CV as an effective signal for monitoring sample diversity during GRPO training.  We use it as a process metric: when Reward CV is low, hyper-parameters (e.g., temperature, clipping coefficient, learning rate, etc.) should be adjusted to increase exploration and improve learning signal quality.
    % \item For complex reasoning tasks such as code generation, reward shaping is critical for effective GRPO optimization. Purely accuracy-based rewards often produce sparse supervision signals, limiting stable policy improvement.

    % \item GRPO is highly sensitive to how reward-shaping hyperparameters are tuned. Increasing the reward scale can speed up optimization, but if reward values become excessively large, training can become unstable and convergence may suffer.
    
    % \item In LoRA-based GRPO fine-tuning, layer-wise contributions are highly non-uniform. Weight displacement patterns vary substantially across tasks and model families, and careful layer selection can improve downstream benchmark performance.
    
    % \item Module-level adaptation in LoRA-GRPO is similarly heterogeneous. Different projection modules exhibit distinct update dynamics, suggesting that selective module adaptation improves optimization efficiency and generalization.
\end{itemize}

We provide a blueprint for the reliable adoption of GRPO in resource-constrained environments. These results suggest that for SLMs to truly excel in agentic and edge AI applications, RFT must move beyond behavioral metrics and toward a deeper understanding of internal representation dynamics. Future work will explore the scaling laws of these mechanistic interventions as SLMs continue to shrink in size but grow in task complexity.

% \section*{References}

\bibliographystyle{plain} 
\bibliography{references}

% References follow the acknowledgments in the camera-ready paper. Use unnumbered first-level heading for
% the references. Any choice of citation style is acceptable as long as you are
% consistent. It is permissible to reduce the font size to \verb+small+ (9 point)
% when listing the references.
% Note that the Reference section does not count towards the page limit.
% \medskip

% {
% \small

% [1] Alexander, J.A.\ \& Mozer, M.C.\ (1995) Template-based algorithms for
% connectionist rule extraction. In G.\ Tesauro, D.S.\ Touretzky and T.K.\ Leen
% (eds.), {\it Advances in Neural Information Processing Systems 7},
% pp.\ 609--616. Cambridge, MA: MIT Press.

% [2] Bower, J.M.\ \& Beeman, D.\ (1995) {\it The Book of GENESIS: Exploring
%   Realistic Neural Models with the GEneral NEural SImulation System.}  New York:
% TELOS/Springer--Verlag.

% [3] Hasselmo, M.E., Schnell, E.\ \& Barkai, E.\ (1995) Dynamics of learning and
% recall at excitatory recurrent synapses and cholinergic modulation in rat
% hippocampal region CA3. {\it Journal of Neuroscience} {\bf 15}(7):5249-5262.
% }

%%%%%%%%%%%%%%%%%%%%%%%%%%%%%%%%%%%%%%%%%%%%%%%%%%%%%%%%%%%%

\appendix
\section{Appendix}

\subsection{GRPO Formulation}\label{app:grpo-equation}

For a prompt \(x \sim \mathcal{D}_{\mathrm{train}}\), let
\(\pi_{\theta}(y \mid x)\) be the trainable policy. At GRPO step \(k\), we
sample a group of completions from the current policy:
\begin{equation}
    y_1,\ldots,y_G
    \sim
    \pi_{\theta_k}(\cdot \mid x),
    \qquad
    y_i = (y_{i,1},\ldots,y_{i,T_i}).
\end{equation}

Each completion receives a scalar reward
\begin{equation}
    r_i
    =
    R_{\lambda}(x,y_i),
    \qquad
    i=1,\ldots,G .
\end{equation}

The group-relative advantage of completion \(y_i\) and policy ratio are respectively,
\begin{equation}
    \hat{A}_i = \frac{r_i - \operatorname{mean}(r_x)}{\operatorname{std}(r_x)}, 
    \qquad 
    \rho_{i,t}(\theta) = \frac{\pi_{\theta}(y_{i,t} \mid x, y_{i,<t})}{\pi_{\theta_k}(y_{i,t} \mid x, y_{i,<t})}
\end{equation}

We penalize deviation from a fixed reference policy
\(\pi_{\refmodel}\). A sample-level KL penalty is
\begin{equation}
    \widehat{K}_{i,t}(\theta)
    =
    \frac{
    \pi_{\refmodel}(y_{i,t}\mid s_{i,t})
    }{
    \pi_{\theta}(y_{i,t}\mid s_{i,t})
    }
    -
    \log
    \frac{
    \pi_{\refmodel}(y_{i,t}\mid s_{i,t})
    }{
    \pi_{\theta}(y_{i,t}\mid s_{i,t})
    }
    -1 .
\end{equation}

The GRPO lower-level objective for fixed design \(\lambda\) is
\begin{equation}
\label{eq:grpo_inner_objective}
\begin{aligned}
    \mathcal{J}_{\mathrm{GRPO}}^{\lambda}
    (\theta;\theta_k)
    =
    \E_{x \sim \mathcal{D}_{\mathrm{train}},
    \{y_i\}_{i=1}^{G} \sim \pi_{\theta_k}}
    \Bigg[
    \frac{1}{G}
    \sum_{i=1}^{G}
    \frac{1}{T_i}
    \sum_{t=1}^{T_i}
    \bigg(
    &
    \min
    \Big[
    \rho_{i,t}(\theta)\widehat{A}_i,
    \clip
    \left(
    \rho_{i,t}(\theta),
    1-\epsilon,
    1+\epsilon
    \right)
    \widehat{A}_i
    \Big]
    \\
    &-
    \beta
    \widehat{K}_{i,t}(\theta)
    \bigg)
    \Bigg].
\end{aligned}
\end{equation}

The inner GRPO training dynamics are
\begin{equation}
\label{eq:grpo_inner_update}
    \theta_{k+1}(\lambda)
    \approx
    \arg\max_{\theta \in \Theta_{\lambda}}
    \mathcal{J}_{\mathrm{GRPO}}^{\lambda}
    \left(
    \theta;\theta_k(\lambda)
    \right),
    \qquad
    k=0,\ldots,T(\lambda)-1,
\end{equation}
with initialization
\begin{equation}
    \theta_0(\lambda)
    =
    \theta_{\refmodel}.
\end{equation}

\subsection{Mechanistic Feedback and Evaluation} \label{app:mech-feedback}

% To constrain output-level drift, define the calibration KL
% \begin{equation}
% \label{eq:calibration_kl}
%     \mathrm{KL}_{\mathcal{C}}
%     \left(
%     \pi_{\theta}
%     \Vert
%     \pi_{\refmodel}
%     \right)
%     =
%     \E_{s \in \mathcal{P}(\mathcal{C})}
%     \left[
%     \KL
%     \left(
%     \pi_{\theta}(\cdot\mid s)
%     \Vert
%     \pi_{\refmodel}(\cdot\mid s)
%     \right)
%     \right],
% \end{equation}
% where \(\mathcal{P}(\mathcal{C})\) denotes the set of token prefixes induced by
% calibration prompts \(\mathcal{C}\).

To estimate weight-level drift, let \(W_g^{\theta}\) denote the weight tensor
of parameter group \(g\) under model \(\theta\). A group \(g\) may correspond
to a layer-module pair, such as an attention projection, MLP projection,
embedding matrix, or language-model head. For a collection of parameter groups
\(\mathcal{G}\), define the normalized weight displacement
\begin{equation}
\label{eq:weight_displacement}
    \WDisp_{\mathcal{G}}
    \left(
    \theta,
    \theta_{\refmodel}
    \right)
    =
    \frac{
    \sqrt{
    \sum_{g \in \mathcal{G}}
    \left\|
    W_g^{\theta}
    -
    W_g^{\theta_{\refmodel}}
    \right\|_F^2
    }
    }{
    \sqrt{
    \sum_{g \in \mathcal{G}}
    \left\|
    W_g^{\theta_{\refmodel}}
    \right\|_F^2
    +
    \delta_w
    }
    },
\end{equation}
where \(\delta_w>0\) is a numerical stabilizer.

For LoRA adaptation, the effective adapted weight can be written as
\begin{equation}
    W_g^{\theta}
    =
    W_g^{\theta_{\refmodel}}
    +
    \frac{\alpha_g}{r_g}
    B_g A_g ,
\end{equation}
where \(r_g\) is the LoRA rank and \(\alpha_g\) is the LoRA scaling parameter.
In this case, the weight displacement becomes
\begin{equation}
\label{eq:lora_weight_displacement}
    \WDisp_{\mathcal{G}}^{\mathrm{LoRA}}
    \left(
    \theta,
    \theta_{\refmodel}
    \right)
    =
    \frac{
    \sqrt{
    \sum_{g \in \mathcal{G}}
    \left\|
    \frac{\alpha_g}{r_g}
    B_g A_g
    \right\|_F^2
    }
    }{
    \sqrt{
    \sum_{g \in \mathcal{G}}
    \left\|
    W_g^{\theta_{\refmodel}}
    \right\|_F^2
    +
    \delta_w
    }
    } .
\end{equation}

To estimate activation-level drift, let
\(h^{\theta}_{\ell,x,t,d}\) denote the hidden activation of model
\(\pi_{\theta}\) at layer \(\ell\), prompt \(x\), token \(t\), and dimension
\(d\). Let \(m_{x,t}\in\{0,1\}\) be a token mask. The normalized activation
shift is
\begin{equation}
\label{eq:activation_shift}
    \Shift_{\mathcal{C},\ell}
    \left(
    \theta,
    \theta_{\refmodel}
    \right)
    =
    \frac{
    \sqrt{
    \sum_{x \in \mathcal{C}}
    \sum_{t=1}^{T_x}
    \sum_{d=1}^{D}
    m_{x,t}
    \left(
    h^{\theta}_{\ell,x,t,d}
    -
    h^{\theta_{\refmodel}}_{\ell,x,t,d}
    \right)^2
    }
    }{
    \sqrt{
    \sum_{x \in \mathcal{C}}
    \sum_{t=1}^{T_x}
    \sum_{d=1}^{D}
    m_{x,t}
    \left(
    h^{\theta_{\refmodel}}_{\ell,x,t,d}
    \right)^2
    +
    \delta_h
    }
    },
\end{equation}
where \(\delta_h>0\) prevents division by zero.

These two metrics weight displacement, $\WDisp_{\mathcal{G}}^{\mathrm{LoRA}}$, and activation shift, $\Shift_{\mathcal{C},\ell}$ provide insight into GRPO training health.

\begin{figure}[t]
    \centering
    \includegraphics[width=\linewidth]{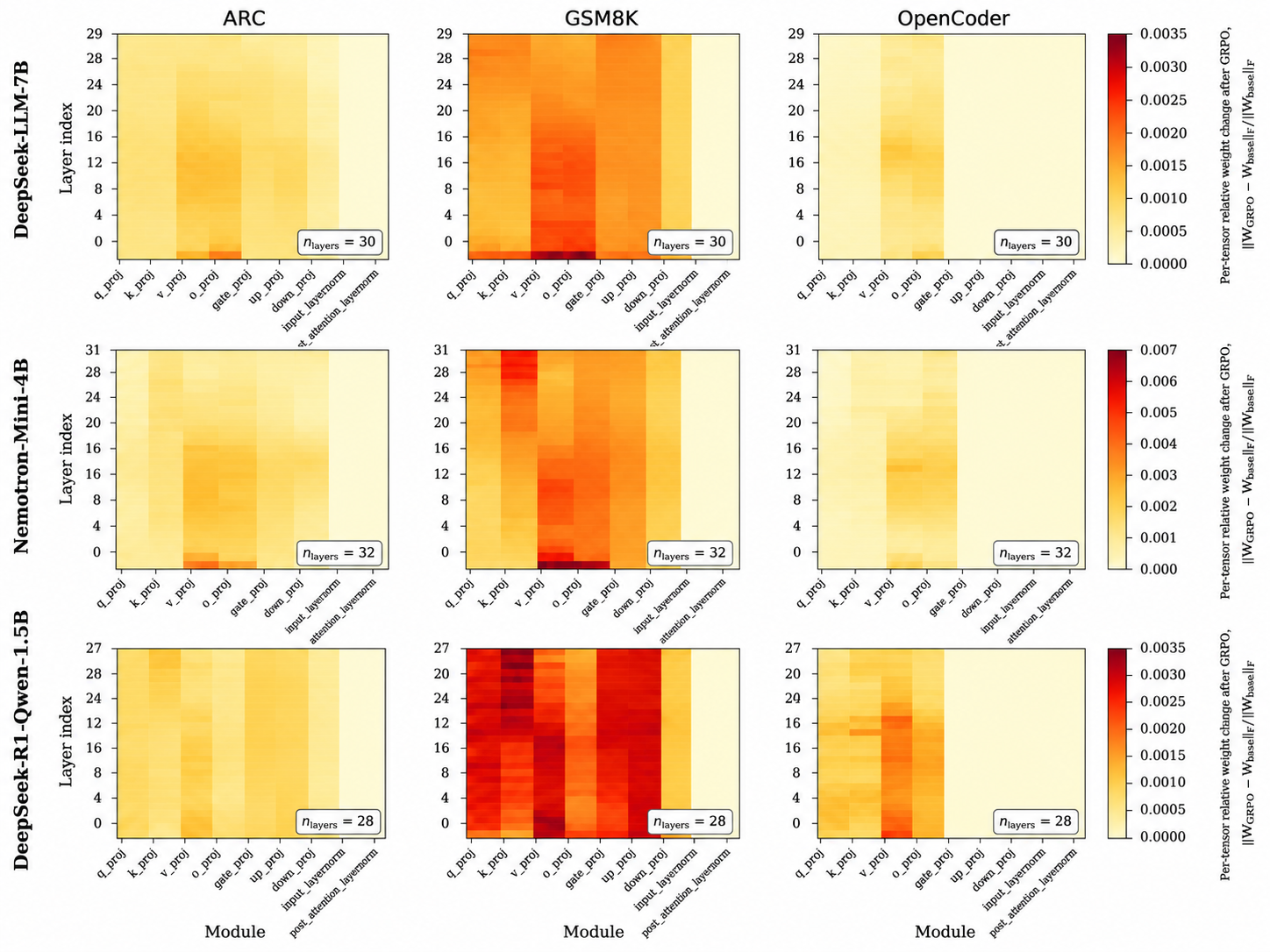}
    \caption{
    Layer-wise parameter displacement after GRPO across model families and tasks. 
    Each heatmap reports the per-tensor relative Frobenius change 
    $\|W_{\mathrm{GRPO}} - W_{\mathrm{base}}\|_F / \|W_{\mathrm{base}}\|_F$ 
    for each module and layer. 
    Rows correspond to model families and columns correspond to reasoning datasets. 
    GRPO induces structured rather than uniform parameter movement: updates are concentrated in attention and MLP projection tensors, while layer-normalization parameters change little. 
    GSM8K produces the largest displacement across models, especially for DeepSeek-R1-Qwen-1.5B, whereas ARC and OpenCoder induce more localized changes.
    }
    \label{fig:weight_delta_heatmap}
\end{figure}
\begin{figure}[H]
    \centering
    \includegraphics[width=\linewidth]{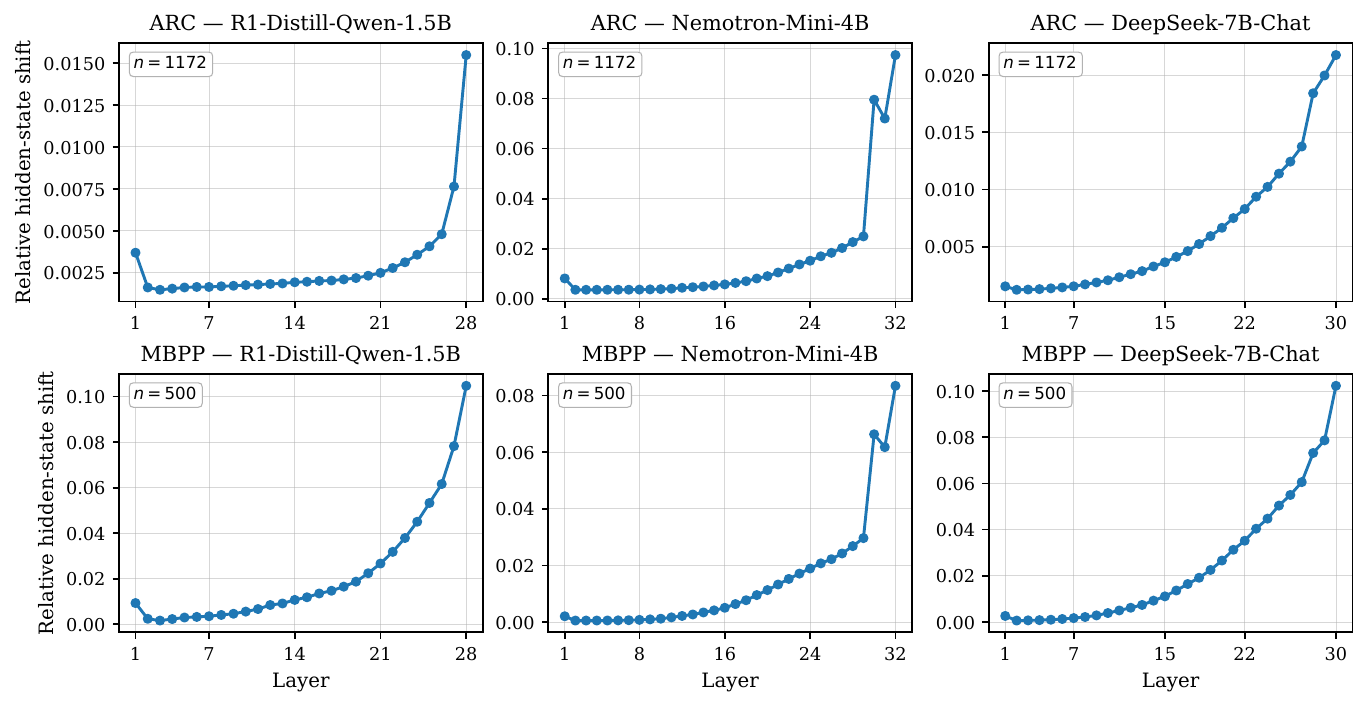}
    \caption{
    Layer-wise relative hidden-state shift induced by GRPO finetuning, base model vs. LoRA-adapted GRPO checkpoint. For
  each prompt $p$ and transformer block $\ell$ (post-block residual stream; the embedding layer is excluded), we compute         
  $s_{p,\ell} = | (\mathbf{H}^{\text{grpo}}{p,\ell} - \mathbf{H}^{\text{base}}{p,\ell}) \odot \mathbf{m}_p |F ,/, |     
  \mathbf{H}^{\text{base}}{p,\ell} \odot \mathbf{m}_p |F$, where $\mathbf{H}{p,\ell} \in \mathbb{R}^{T \times d}$ is the         
  layer-$\ell$ hidden state of prompt $p$, $\mathbf{m}p$ is the attention mask zeroing padding positions, and $|\cdot|F$ is the 
  Frobenius norm. Each panel reports the per-layer mean $\bar{s}\ell = \tfrac{1}{N}\sum_p s{p,\ell}$ over the dataset's full    
  prompt set ($N{=}1172$ for AI2-ARC test, $N{=}500$ for MBPP test). Rows: evaluation dataset (top: AI2-ARC, bottom: MBPP). 
  Columns: base model (left to right: DeepSeek-R1-Distill-Qwen-1.5B, Nemotron-Mini-4B-Instruct, DeepSeek-LLM-7B-Chat). Across all
   six combinations the shift is small and roughly flat through the bulk of the network and grows sharply in the last 2–3 layers,
   indicating that GRPO concentrates its representational change near the output, leaving early- and mid-stack features almost   
  unchanged. 
    }
    \label{fig:logit_shift}
\end{figure}

\begin{figure}[H]
\centering
\begin{subfigure}{\linewidth}
    \centering
    \includegraphics[width=\linewidth]{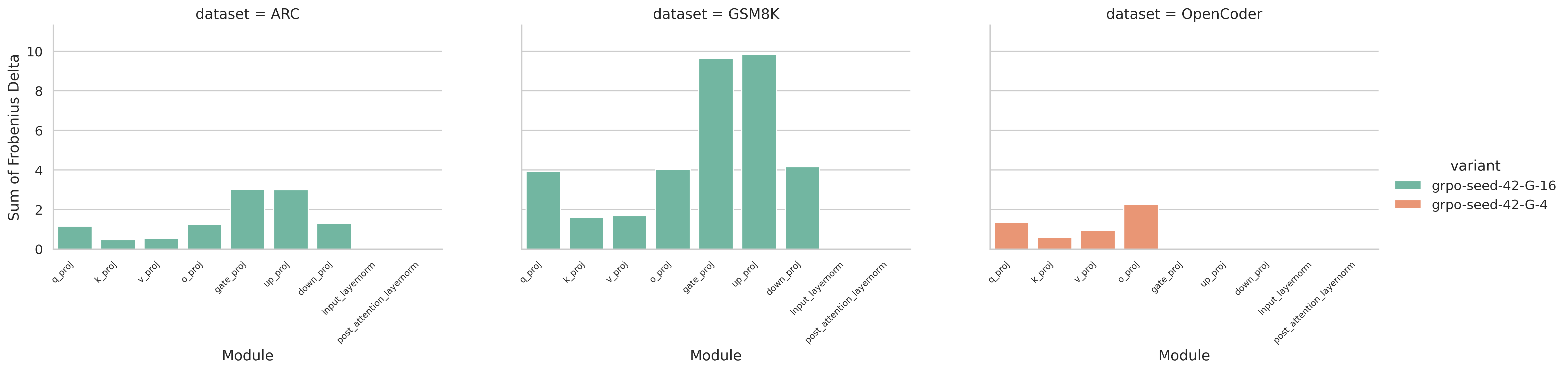}
    \caption{DeepSeek-R1-Qwen-1.5}
\end{subfigure}
\vspace{2pt}
\begin{subfigure}{\linewidth}
    \centering
    \includegraphics[width=\linewidth]{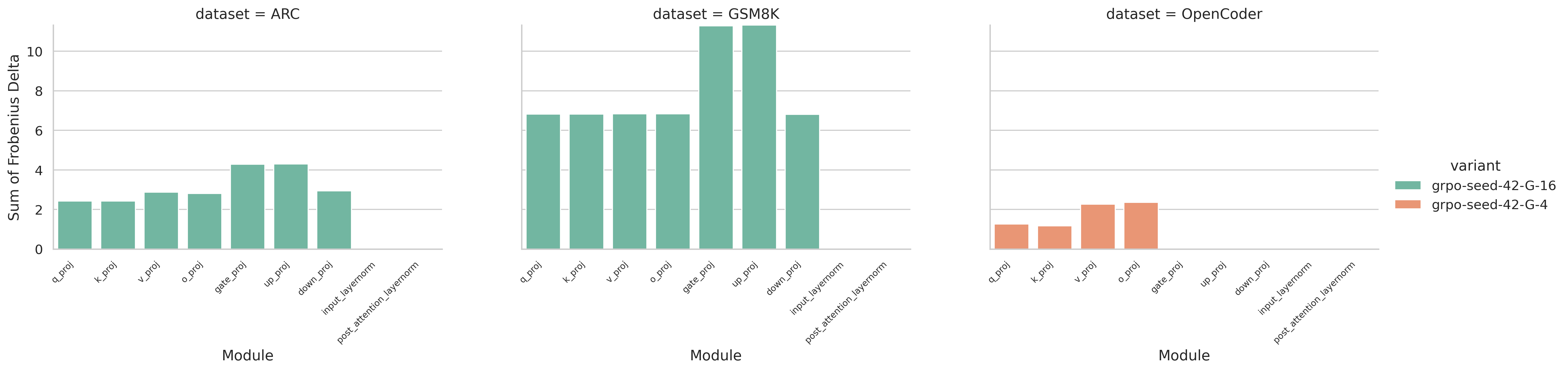}
    \caption{DeepSeek-LLM-7B-chat}
\end{subfigure}
\vspace{2pt}
\begin{subfigure}{\linewidth}
    \centering
    \includegraphics[width=\linewidth]{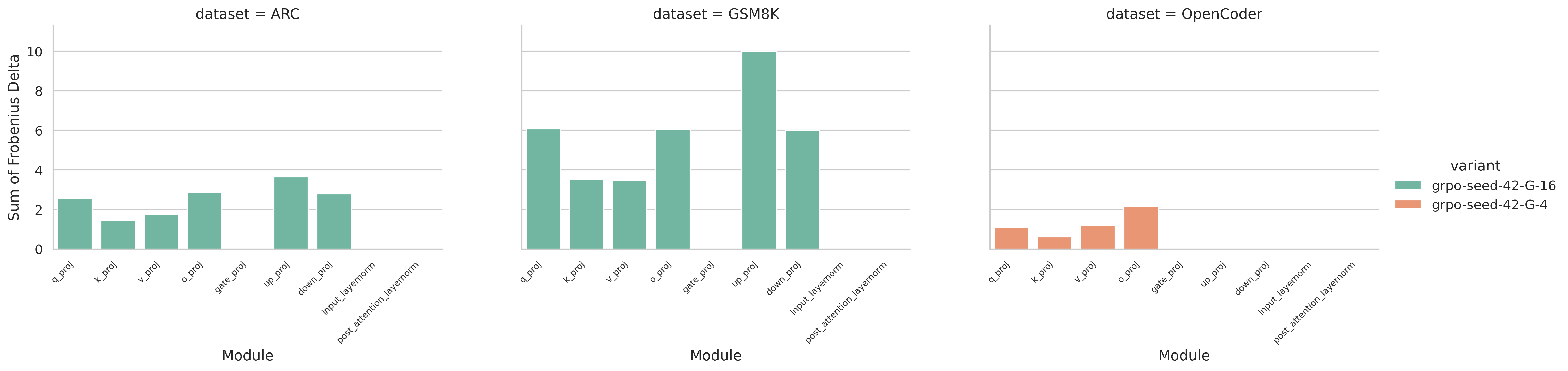}
    \caption{Nemotron4B}
\end{subfigure}

\caption{\textbf{Module-wise weight update mass across models.} Each subplot shows how GRPO-induced updates are distributed across transformer modules, revealing consistent patterns in specific blocks across architectures.}
\label{fig:module_update_mass_all}
\end{figure}

Figure~\ref{fig:weight_delta_heatmap} visualizes the relative parameter displacement induced by GRPO across model families, tasks, layers, and modules. 
Each heatmap reports the per-tensor relative Frobenius change 
$\|W_{\mathrm{GRPO}} - W_{\mathrm{base}}\|_F / \|W_{\mathrm{base}}\|_F$ 
for a given module and layer, with rows corresponding to model families and columns corresponding to ARC, GSM8K, and OpenCoder. 
Across all models, the largest changes are concentrated in attention and MLP projection matrices, while layer-normalization parameters exhibit comparatively small movement. 
The magnitude and localization of these updates are strongly task-dependent: GSM8K induces substantially larger parameter displacement than ARC or OpenCoder, especially in the projection modules.

The pattern is also model-dependent. 
DeepSeek-R1-Qwen-1.5B exhibits the broadest and most intense weight changes on GSM8K, with high relative movement across several projection modules and many layers. 
Nemotron-Mini-4B shows a similarly task-sensitive pattern but with a larger overall scale, particularly on GSM8K, as reflected by its wider color range. 
DeepSeek-LLM-7B exhibits more moderate changes, with GSM8K again producing the strongest displacement, but ARC and OpenCoder remaining comparatively localized. 
Overall, the figure shows that GRPO does not update the model uniformly: parameter movement is structured by task, architecture, module type, and layer depth, suggesting that the optimization primarily adapts a subset of projection tensors rather than producing diffuse global drift.

\subsection{Reward Shaping Scores}

% % \begin{table}[!htbp]
% \begin{table}[h]
% \centering
% \caption{Reward shaping components used in GRPO training.}
% \label{tab:reward_config}
% \small % Slightly smaller text to fit better in two-column layouts if needed
% \begin{tabular}{lr} % Use 'r' for values to align decimal points/integers
% \toprule
% \textbf{Benchmark} & \textbf{Value} \\
% \midrule
% \rowcolor{gray!10} \multicolumn{2}{l}{\textbf{GSM8K}} \\ % Light shading for categories
% Correctness Reward            & 2.0   \\
% XML Count Reward   & 0.5   \\
% Numeric Answer Present Reward & 0.5 \\
% Format Reward & 0.5 \\
% Reasoning Present Reward       & 0.1   \\
% Answer Present Reward       & 0.1   \\
% \addlinespace[0.5em] % Adds subtle vertical breathing room

% \rowcolor{gray!10} \multicolumn{2}{l}{\textbf{ARC}} \\
% Correctness Reward               & 1.0   \\
% Valid Option Reward        & 0.5  \\
% XML Count Reward               & 0.5 \\
% Format Reward               & 0.5 \\
% \addlinespace[0.5em]

% \rowcolor{gray!10} \multicolumn{2}{l}{\textbf{OpenCoder}} \\
% Correctness Reward        & 1.0 \\
% Code Complexity Reward               & 1.0 \\
% XML Count Reward               & 0.5 \\
% Syntax Reward        & 0.5 \\
% Reasoning Present Reward               & 0.1 \\
% \bottomrule
% \end{tabular}
% \end{table}

\begin{table}[h]
\centering
\caption{Reward shaping components used in GRPO training.}
\label{tab:reward_config}
\small
\setlength{\tabcolsep}{4pt}
\begin{tabular}{lrl}
\toprule
\textbf{Component} & \textbf{Value} & \textbf{Behavior} \\
\midrule
\rowcolor{gray!10} \multicolumn{3}{l}{\textbf{GSM8K}} \\
Correctness Reward            & 2.0 & Final answer matches ground truth \\
XML Count Reward              & 0.5 & All required XML tags and sections are present \\
Numeric Answer Present Reward & 0.5 & Response contains a numeric answer \\
Format Reward                 & 0.5 & Output follows expected structure \\
Reasoning Present Reward      & 0.1 & Reasoning block is non-empty \\
Answer Present Reward         & 0.1 & Answer block is non-empty \\
\addlinespace[0.5em]
\rowcolor{gray!10} \multicolumn{3}{l}{\textbf{ARC}} \\
Correctness Reward            & 2.0 & Selected option matches ground truth \\
Valid Option Reward           & 0.5 & Answer is one of the valid choices \\
XML Count Reward              & 0.5 & All required XML tags and sections present \\
Format Reward                 & 0.5 & Output follows expected structure \\
\addlinespace[0.5em]
\rowcolor{gray!10} \multicolumn{3}{l}{\textbf{OpenCoder}} \\
Correctness Reward            & 1.0 & Code passes all test cases \\
Code Complexity Reward        & 1.0 & Penalizes trivial or degenerate solutions \\
XML Count Reward              & 0.5 & All required XML tags and sections present \\
Syntax Reward                 & 0.5 & Generated code parses without errors \\
Reasoning Present Reward      & 0.1 & Reasoning block is non-empty \\
\bottomrule
\end{tabular}
\end{table}\label{app:rewards}
\FloatBarrier

\subsection{Training Hyperparameters}\label{app:params}
\begin{table}[h]
\centering
\caption{GRPO training configurations across GSM8K, OpenCoder, and ARC-Challenge.}
\label{tab:all_configs}
\scriptsize
\renewcommand{\arraystretch}{1.2}

\begin{tabular}{l c c c}
\toprule
\textbf{Category} & \textbf{GSM8K} & \textbf{OpenCoder} & \textbf{ARC-Challenge} \\
\midrule

\multicolumn{4}{l}{\textit{Training}} \\
Learning Rate & $1\times10^{-5}$ & $5\times10^{-6}$ & $1\times10^{-5}$ \\
Epochs & 2 & 1 & 2 \\
Batch Size & 4 & 1 & 4 \\
Gradient Accumulation & 4 & 64 & 4 \\
Temperature & 0.8 & 0.8 & 0.8 \\
Max Prompt Length & 256 & 256 & 256 \\
Max Completion Length & 512 & 512 & 512 \\
Eval Steps & 100 & 100 & 100 \\
KL Coefficient ($\beta$) & 0.005 & 0.005 & 0.003 \\
Clipping Coefficient ($\epsilon$) & 0.2 & 0.2 & 0.2 \\
Train/Test Split & train/test & train[95\% split] & train/validation \\
\midrule

\multicolumn{4}{l}{\textit{LoRA}} \\
Use LoRA & True & True & True \\
Rank ($r$) & 16 & 8 & 16 \\
Alpha & 64 & 16 & 64 \\
Dropout & 0.05 & 0.1 & 0.05 \\
Target Modules & q,k,v,o,up,down,gate & q,k,v,o & q,k,v,o,up,down,gate \\
\midrule

\multicolumn{4}{l}{\textit{Optimizer}} \\
Optimizer & adamw\_8bit & adamw\_8bit & adamw\_8bit \\
Adam $\beta_1$ & 0.9 & 0.9 & 0.9 \\
Adam $\beta_2$ & 0.99 & 0.99 & 0.99 \\
Weight Decay & 0.1 & 0.1 & 0.1 \\
Warmup Ratio & 0.1 & 0.1 & 0.1 \\
LR Scheduler & cosine & cosine & cosine \\
Max Grad Norm & 0.1 & 0.1 & 0.1 \\
\bottomrule
\end{tabular}
\end{table}

\newcommand{\cmark}{\textcolor{green!55!black}{\ding{51}}}
\newcommand{\xmark}{\textcolor{red!75!black}{\ding{55}}}

\subsection{Qualitative Comparison: Base vs.\ GRPO Predictions}
\label{app:qualitative}
\subsubsection{GSM8K Dataset}
\label{appendix:gsm}
We provide two representative GSM8K samples that illustrate \emph{how}
GRPO fine-tuning improves over the base checkpoint. Sample~1 shows a
correction of an arithmetic slip in Deepseek-LLM-7B; Sample~2 shows GRPO
fixing a final-answer formatting failure in Nemotron-Mini-4B.

% --------------------------------------------------------------------------
% Sample 1: DS-LLM-7B  --  Josh flips a house
% --------------------------------------------------------------------------
\begin{table}[H]
\centering
\caption{GSM8K sample where GRPO fine-tuning corrects an arithmetic
error made by the base \textbf{Deepseek-LLM-7B-Chat}. Both GRPO seeds (G16,
$s\!=\!42$ and $s\!=\!3407$) recover the correct answer with the same
explicit decomposition.}
\label{tab:qual_ds7b}
\small
\renewcommand{\arraystretch}{1.2}
\begin{tabular}{p{0.46\linewidth} p{0.46\linewidth}}
\toprule
\multicolumn{2}{p{0.94\linewidth}}{\textbf{Question.} Josh decides to
try flipping a house. He buys a house for \$80{,}000 and then puts in
\$50{,}000 in repairs. This increased the value of the house by 150\%.
How much profit did he make? \newline\textbf{Ground truth: \$70{,}000}}\\
\midrule
\textbf{Deepseek-LLM-7B-Chat (base)} \xmark &
\textbf{Deepseek-LLM-7B GRPO} \cmark \\
Predicted: \textbf{\$150{,}000} &
Predicted: \textbf{\$70{,}000} \\
\midrule
{\scriptsize\ttfamily
Josh bought the house for \$80{,}000 and invested \$50{,}000 in
repairs, so the total cost was \$80{,}000 + \$50{,}000 = \$130{,}000.\newline
The repairs increased the value of the house by 150\%, so the new
value of the house is \$80{,}000 + 1.5 $\times$ \$80{,}000 =
\textbf{\$280{,}000}.\newline
To calculate the profit, we subtract the total cost from the new value
of the house, so \$280{,}000 $-$ \$130{,}000 = \$150{,}000.\newline
Therefore, Josh made a profit of \$150{,}000.\newline
The answer is: 150000} &
{\scriptsize\ttfamily
Josh bought the house for \$80{,}000.\newline
He spent \$50{,}000 on repairs.\newline
The total cost was \$80{,}000 + \$50{,}000 = \$130{,}000.\newline
After the repairs, the value of the house increased by 150\%, so the
new value is \$80{,}000 + (\$80{,}000 $\times$ 1.5) = \$80{,}000 +
\$120{,}000 = \$200{,}000.\newline
The profit Josh made is the new value of the house minus the cost, so
it is \$200{,}000 $-$ \$130{,}000 = \$70{,}000.\newline
The answer is: 70000} \\
\bottomrule
\end{tabular}
\end{table}
\FloatBarrier
\noindent\textbf{Observation.} The base model selects the correct
high-level plan but commits an arithmetic slip when summing
\$80{,}000\,+\,\$120{,}000. The GRPO checkpoint inserts an explicit
intermediate sum before subtracting, eliminating the error.

% --------------------------------------------------------------------------
% Sample 2: Nemotron-Mini-4B  --  Janet's ducks
% --------------------------------------------------------------------------
\begin{table}[!ht]
\centering
\caption{GSM8K sample where GRPO fine-tuning fixes a final-answer
formatting failure in \textbf{Nemotron-Mini-4B}. The base reaches the correct
intermediate value (\$18) but emits a malformed final token
\texttt{1\}}; the GRPO checkpoint produces a clean final answer.}
\label{tab:qual_nemotron4b}
\small
\renewcommand{\arraystretch}{1.2}
\begin{tabular}{p{0.46\linewidth} p{0.46\linewidth}}
\toprule
\multicolumn{2}{p{0.94\linewidth}}{\textbf{Question.} Janet's ducks lay
16 eggs per day. She eats three for breakfast every morning and bakes
muffins for her friends every day with four. She sells the remainder
at the farmers' market daily for \$2 per fresh duck egg. How much in
dollars does she make every day at the farmers' market?
\newline\textbf{Ground truth: \$18}}\\
\midrule
\textbf{Nemotron-4B (base)} \xmark &
\textbf{Nemotron-4B GRPO} \cmark \\
Predicted: \textbf{1} (malformed) &
Predicted: \textbf{18} \\
\midrule
{\scriptsize\ttfamily
The ducks lay 16 eggs per day.\newline
Janet eats 3 eggs per day so the number of eggs available to her is
16 $-$ 3 = 13 eggs.\newline
She bakes muffins with 4 eggs, so the number of eggs left is
13 $-$ 4 = 9 eggs.\newline
Out of these 9 eggs, she sells 9 $\times$ 2 = 18 dollars worth of
eggs at the farmers' market.\newline
Thus she makes \textbf{1\}} dollars every day at the farmers' market.} &
{\scriptsize\ttfamily
Sure! Let's break down the problem step by step.\newline
1. Janet's ducks lay 16 eggs per day.\newline
2. She eats 3 eggs for breakfast every morning, so she eats 3 eggs
per day.\newline
3. She bakes muffins for her friends every day with 4 eggs. So she
uses 4 eggs per day for baking.\newline
4. In total, she uses 3 + 4 = 7 eggs per day.\newline
5. She has 16 $-$ 7 = 9 eggs left to sell at the farmers' market.\newline
6. She sells 9 eggs at \$2 per egg, so she makes 9 $\times$ 2 = \$18
at the farmers' market per day.\newline
Therefore, the answer is \textbf{\$18}.} \\
\bottomrule
\end{tabular}
\end{table}

\noindent\textbf{Observation.} The base model performs the arithmetic
correctly but truncates the final answer to the malformed token
\texttt{1\}}, which the answer extractor reads as ``1''. GRPO
fine-tuning eliminates this failure mode and emits a well-formed
\texttt{\$18}.

% --------------------------------------------------------------------------
% Sample 3: DeepSeek-R1-Qwen-1.5B  --  Wendi's chickens
% --------------------------------------------------------------------------
\begin{table}[!ht]
\centering
\caption{GSM8K sample where GRPO fine-tuning corrects a
self-doubt / reasoning-loop failure in
\textbf{DeepSeek-R1-Distill-Qwen-1.5B}. The base checkpoint reaches
the correct intermediate answer (20) but oscillates between two
interpretations of ``three cups'' and ultimately commits to the wrong
final value. The GRPO checkpoint (G4, $s\!=\!42$) executes a single
clean budget calculation.}
\label{tab:qual_dsqwen15b}
\small
\renewcommand{\arraystretch}{1.2}
\begin{tabular}{p{0.46\linewidth} p{0.46\linewidth}}
\toprule
\multicolumn{2}{p{0.94\linewidth}}{\textbf{Question.} Every day,
Wendi feeds each of her chickens three cups of mixed chicken feed,
containing seeds, mealworms and vegetables to help keep them
healthy. She gives the chickens their feed in three separate meals.
In the morning, she gives her flock of chickens 15 cups of feed. In
the afternoon, she gives her chickens another 25 cups of feed. How
many cups of feed does she need to give her chickens in the final
meal of the day if the size of Wendi's flock is 20 chickens?
\newline\textbf{Ground truth: 20}}\\
\midrule
\textbf{DS-R1-Qwen-1.5B (base)} \xmark &
\textbf{DS-R1-Qwen-1.5B GRPO} (G4, $s\!=\!42$) \cmark \\
Predicted: \textbf{25} &
Predicted: \textbf{20} \\
\midrule
{\scriptsize\ttfamily
Okay, so I need to figure out how many cups of feed Wendi needs to
give her 20 chickens in the final meal of the day.\newline
Each chicken gets three cups per day, split into three meals. So
total feed per day is $3 \times 20 = 60$ cups.\newline
She already gave $15 + 25 = 40$ cups, so the remaining is
$60 - 40 = \textbf{20}$ cups.\newline
\textit{Alternatively, maybe she feeds each chicken 3 cups every
day, but the 3 cups are per meal, not per chicken.}\newline
If each meal is 1 cup per chicken, then for 20 chickens each meal is
20 cups. Morning: gave 15, short by 5. Afternoon: gave 25, over by
5\ldots\newline
\textit{Wait, this is getting confusing.}\newline
Maybe she can't give 5 cups over, so she can only give 20 cups in
the afternoon, but the problem says she gave 25\ldots\newline
[oscillates between the two readings for several more paragraphs]\newline
The answer is: \textbf{25}} &
{\scriptsize\ttfamily
Wendi feeds her 20 chickens three cups of mixed feed each day,
divided into three meals.\newline
Total feed required per day:
$3 \text{ cups/chicken/day} \times 20 \text{ chickens}
= 60 \text{ cups/day}$.\newline
Total feed given in the morning and afternoon:
$15 + 25 = 40 \text{ cups}$.\newline
Subtract the feed already given from the total daily requirement:
$60 - 40 = 20 \text{ cups}$.\newline
Thus, the final meal of the day requires
\textbf{\boxed{20}} cups of feed.} \\
\bottomrule
\end{tabular}
\end{table}
\FloatBarrier
\noindent\textbf{Observation.} Unlike Samples~1 and~2, where the base
model fails through a single arithmetic slip or a malformed final
token, here the base failure is qualitatively different: the model
\emph{derives the correct answer} (20) early in its chain of thought,
then re-interprets the problem statement (``three cups'' as
per-meal rather than per-day), produces a contradictory per-meal
accounting, and oscillates between the two readings for the
remainder of its trace before committing to the afternoon quantity
(25) as the final answer. The GRPO checkpoint, by contrast, commits
to a single decomposition --- daily budget minus feed already
delivered --- and terminates as soon as that calculation is complete.
This illustrates a third mode of GRPO improvement on GSM8K beyond
arithmetic correction (Sample~1) and format repair (Sample~2):
\textbf{suppression of self-doubt loops}, in which the verifiable
reward signal pushes the policy away from long, vacillating chains
of thought toward shorter, committed reasoning that retains the
first correct derivation.
\subsubsection{ARC Dataset}
\label{appendix:arc}

 We present representative ARC examples to highlight differences between base and GRPO-tuned models on conceptual and reasoning-intensive questions. The comparisons focus on common failure modes in base models, including conceptual confusion, shallow reasoning, and reliance on surface-level associations. GRPO fine-tuning improves performance by promoting better conceptual grounding, clearer reasoning structure, and more accurate alignment between question semantics and final predictions.
 
 \begin{table}[H]
\centering
\caption{ARC sample where GRPO fine-tuning corrects a conceptual
confusion between rotation and revolution. The base model conflates
Earth's orbital period with its rotational period, while the GRPO
checkpoint correctly identifies axial rotation.}
\label{tab:qual_arc_rotation_verbatim}
\small
\renewcommand{\arraystretch}{1.2}
\begin{tabular}{p{0.46\linewidth} p{0.46\linewidth}}
\toprule
\multicolumn{2}{p{0.94\linewidth}}{\textbf{Question.} The period of rotation for each planet in our solar system is different. The period of rotation for Earth is:\newline
\centerline{(A) 24 hours \quad (B) 28 days \quad (C) 13 weeks \quad (D) 1 year} \newline\textbf{Ground truth: (A)}}\\
\midrule
\textbf{DS-R1-Qwen-1.5B (base)} \xmark &
\textbf{DS-R1-Qwen-1.5B GRPO} \cmark \\
Predicted: \textbf{(D)} &
Predicted: \textbf{(A)} \\
\midrule
{\scriptsize\ttfamily
To determine the period of rotation for Earth, I need to recall the definition of a year. A year is the time it takes for Earth to complete one full orbit around the Sun
Next, I should consider the options provided. Option A is 24 hours, which is the duration of a day, not a year. Option B is 28 days, which is approximately the length of a month, not a year. Option C is 13 weeks, which is about the duration of a month as well. Option D is 1 year, which aligns perfectly with the definition of a year.

Therefore, the correct answer is D, 1 year.

\texttt{\textless/think\textgreater}

To determine the period of rotation for Earth, let's analyze the options provided:

\textbf{Choices:}
- \textbf{(A) 24 hours}
- \textbf{(B) 28 days}
- \textbf{(C) 13 weeks}
- \textbf{(D) 1 year}

\textbf{Step-by-Step Explanation:}

1. \textbf{Understanding the Term "Period of Rotation":}
   - The period of rotation refers to the time it takes for a celestial body to complete one full orbit around its axis.

2. \textbf{Definition of a Year:}
   - A \textbf{year} is the time it takes for Earth to complete one full orbit around the Sun.

3. \textbf{Analyzing the Options:}
   - \textbf{Option A (24 hours):} This is the duration of a day, not a year.
   - \textbf{Option B (28 days):} This is approximately the length of a month, not a year.
   - \textbf{Option C (13 weeks):} This is about the duration of a month as well.
   - \textbf{Option D (1 year):} This matches the definition of a year, as it represents Earth's orbital period around the Sun.

\textbf{Conclusion:}

The correct answer is \textbf{(D) 1 year}, as it accurately describes the period of rotation for Earth.

\boxed{D}} &
{\scriptsize\ttfamily
To determine the period of rotation for Earth, I need to recall the definition of a day. A day is the time it takes for Earth to complete one full rotation on its axis, which is 24 hours.

Next, I should consider the options provided. Option A is 24 hours, which directly matches the definition of a day. Option B is 28 days, which is longer than a year and doesn't represent Earth's rotation period. Option C is 13 weeks, which is approximately 91 days, also not matching Earth's rotation. Option D is 1 year, which is much longer than Earth's rotation period.

Therefore, the correct answer is A, 24 hours.

\texttt{\textless/think\textgreater}

To determine the period of rotation for Earth, let's analyze the options step by step.

1. \textbf{Understanding the Question:}
   - The question asks for the period of rotation of Earth.
   - The options are:
     - (A) 24 hours
     - (B) 28 days
     - (C) 13 weeks
     - (D) 1 year

2. \textbf{Recalling the Definition of a Day:}
   - A \textbf{day} is defined as the time it takes for Earth to complete one full rotation on its axis.
   - This duration is exactly \textbf{24 hours}.

3. \textbf{Evaluating the Options:}
   - \textbf{Option A (24 hours):} This matches the definition of a day, making it the correct answer.
   - \textbf{Option B (28 days):} This is longer than a year and does not represent Earth's rotation period.
   - \textbf{Option C (13 weeks):} This is approximately 91 days, which is not the period of Earth's rotation.
   - \textbf{Option D (1 year):} This is much longer than Earth's rotation period.

4. \textbf{Conclusion:}
   - The only option that accurately represents Earth's period of rotation is \textbf{24 hours}.

\boxed{A}} \\
\bottomrule
\end{tabular}
\end{table}
\noindent\textbf{Observation.} The base model conflates rotation with revolution, selecting the Earth's orbital period (1 year) instead of its rotational period. The GRPO checkpoint correctly grounds rotation as axial motion (a day), demonstrating improved conceptual disambiguation and alignment with the question's intent.

\begin{table}[h]
\centering
\caption{ARC sample where GRPO fine-tuning corrects a reasoning ambiguity. The base model misattributes evidence to water presence, while the GRPO checkpoint correctly identifies climatic implication.}
\label{tab:qual_arc_petrified_palm_grpo}
\small
\renewcommand{\arraystretch}{1.2}
\begin{tabular}{p{0.46\linewidth} p{0.46\linewidth}}
\toprule
\multicolumn{2}{p{0.94\linewidth}}{\textbf{Question.} Petrified palm trees are found in sedimentary rock near glaciers. The presence of the petrified palm trees most likely provides evidence for which statement?\newline
\centerline{(A) There was once more water in the area. (B) The area was once grassland.}\newline\centerline{ (C) The climate in the area was once tropical. (D) There are active faults in the area.} \newline \textbf{Ground truth: (C)}}\\
\midrule
\textbf{DeepSeek-LLM-7B-Chat (base)} \xmark &
\textbf{GRPO model} \cmark \\
Predicted: \textbf{(A)} &
Predicted: \textbf{(C)} \\
\midrule
{\scriptsize\ttfamily
The correct answer is (A) There was once more water in the area. The presence of petrified palm trees near glaciers suggests that the area was once near a tropical forest. This would indicate that there was once more water in the area, which is conducive to the growth of palm trees. ... (C) The climate in the area was likely not tropical at the time the palm trees were alive, as they are now fossilized.} &
{\scriptsize\ttfamily
The presence of petrified palm trees in the vicinity of a glacier indicates that the area was once warmer and wetter, which is consistent with a tropical climate. This supports the statement that the area was once tropical. Therefore, the correct answer is (C) The climate in the area was once tropical.} \\
\bottomrule
\end{tabular}
\end{table}
\FloatBarrier
\noindent\textbf{Observation.} The base model outputs only a final label without any supporting reasoning, selecting option (A) based on a superficial association with “support.” In contrast, the GRPO checkpoint correctly identifies the stem as a transport mechanism and maps it to an elevator, which performs an analogous function. This reflects improved abstraction over functional similarity rather than surface-level matching.

\begin{table}[h]
\centering
\caption{ARC sample where GRPO fine-tuning improves analogical reasoning by correctly identifying functional similarity rather than superficial association. The base model produces a shallow incorrect choice with no reasoning, while the GRPO checkpoint identifies the correct transport-function mapping with structured explanation.}
\label{tab:qual_arc_stem_analogy}
\small
\renewcommand{\arraystretch}{1.2}
\begin{tabular}{p{0.46\linewidth} p{0.46\linewidth}}
\toprule
\multicolumn{2}{p{0.94\linewidth}}{\textbf{Question.} The stem is an important part of many plants. Which of the following is most similar to the role performed by the stem of a plant?\newline
\centerline{(A) an anchor holding a boat in place (B) a snack company producing energy bars}\newline\centerline{(C) a sign attracting people (D) an elevator transporting supplies from one floor to another.} \newline \textbf{Ground truth: (D)}}\\
\midrule
\textbf{Nemotron-Mini-4B (base)} \xmark &
\textbf{GRPO model} \cmark \\
Predicted: \textbf{(A)} &
Predicted: \textbf{(D)} \\
\midrule
{\scriptsize\ttfamily
The answer is A) an anchor holding a boat in place} &
{\scriptsize\ttfamily
The answer is (D) an elevator transporting supplies from one floor to another

The stem of a plant performs a similar function as an elevator in a building. It transports materials (water and nutrients) from the roots to the leaves and other parts of the plant, just like how an elevator carries people and supplies from one floor to another.

Here's a brief comparison:

- Stem (plant): Transports water and nutrients from roots to leaves and other parts of the plant.
- Elevator (building): Transports people and supplies from one floor to another.

Option (A) is incorrect because an anchor holds a boat in place, not performs a transport function.

Option (B) is incorrect because a snack company does not produce materials.

Option (C) is incorrect because a sign does not transport materials or people.} \\
\bottomrule
\end{tabular}
\end{table}

\noindent\textbf{Observation.} The base model selects option (A), attributing petrified palm trees primarily to historical water presence rather than climatic conditions, reflecting a weaker distinction between hydrological and climatic inference. In contrast, the GRPO checkpoint correctly associates petrified palm trees near glaciers with a past tropical climate and selects option (C). This demonstrates improved reasoning alignment in distinguishing environmental causality from proxy evidence.
\FloatBarrier

\subsubsection{MBPP}\label{apps:mbpp_examples}

%% ============================================================================
%% MBPP TABLES
%% ============================================================================

\begin{table}[H]
\centering
\caption{MBPP Task 23 sample comparing Nemotron-Mini-4B base and GRPO models.}
\label{tab:mbpp_nemotron}
\small
\renewcommand{\arraystretch}{1.2}
\begin{tabular}{p{0.46\linewidth} p{0.46\linewidth}}
\toprule
\multicolumn{2}{p{0.94\linewidth}}{\textbf{Question.} Write a python function to find the maximum sum of elements of list in a list of lists.}\\
\midrule
\textbf{Nemotron Base} \xmark &
\textbf{Nemotron GRPO} \cmark \\
Pass Rate: \textbf{0/10} &
Pass Rate: \textbf{7/10} \\
\midrule
{\scriptsize\ttfamily
$<$reasoning$>$\newline
To find the maximum sum of elements in a list of lists, we can iterate through each sublist and find the maximum sum for that sublist. We can do this by iterating through each element in the sublist and adding it to a running total, and then updating the maximum sum if necessary.\newline
$<$/reasoning$>$\newline
$<$answer$>$\newline
def maximum\_Sum(matrix):\newline
\hspace*{1em}max\_sum = 0\newline
\hspace*{1em}for sublist in matrix:\newline
\hspace*{2em}current\_sum = 0\newline
\hspace*{2em}for num in sublist:\newline
\hspace*{3em}current\_sum += num\newline
\hspace*{3em}if current\_sum $>$ max\_sum:\newline
\hspace*{4em}max\_sum = current\_sum\newline
\hspace*{1em}return max\_sum\newline
$<$/answer$>$} &
{\scriptsize\ttfamily
$<$reasoning$>$\newline
This problem can be solved using a dictionary to store the elements and their indices. If an element is already in the dictionary, it means it has been seen before, so we return its index. If not, we add it to the dictionary and return -1.\newline
$<$/reasoning$>$\newline
$<$answer$>$\newline
def find\_first\_duplicate(arr):\newline
\hspace*{1em}return next((i for i, x in enumerate(arr) if arr.count(x) $>$ 1), -1)\newline
$<$/answer$>$} \\
\bottomrule
\end{tabular}
\end{table}
\noindent\textbf{Observation.} The base model produces an incorrect logic loop, while the GRPO checkpoint passes the evaluation.

\begin{table}[H]
\centering
\caption{MBPP Task 445 sample comparing Deepseek 7B base and GRPO models.}
\label{tab:mbpp_deepseek7b}
\small
\renewcommand{\arraystretch}{1.2}
\begin{tabular}{p{0.46\linewidth} p{0.46\linewidth}}
\toprule
\multicolumn{2}{p{0.94\linewidth}}{\textbf{Question.} Write a function to perform index wise multiplication of tuple elements in the given two tuples.}\\
\midrule
\textbf{Deepseek 7B Base} \xmark &
\textbf{Deepseek 7B GRPO} \cmark \\
Pass Rate: \textbf{0/10} &
Pass Rate: \textbf{10/10} \\
\midrule
{\scriptsize\ttfamily
$<$reasoning$>$\newline
The function takes two tuples as input and returns a new tuple where the i-th element is the product of the i-th elements of the two input tuples.\newline
$<$/reasoning$>$\newline
$<$answer$>$\newline
def index\_multiplication(t1, t2):\newline
\hspace*{1em}result = []\newline
\hspace*{1em}for i in range(len(t1)):\newline
\hspace*{2em}result.append(t1[i] * t2[i])\newline
\hspace*{1em}return tuple(result)\newline
$<$/answer$>$} &
{\scriptsize\ttfamily
$<$reasoning$>$\newline
The index\_multiplication function takes two tuples as input and returns a new tuple with the index-wise multiplication of the elements of the input tuples.\newline
$<$/reasoning$>$\newline
$<$answer$>$\newline
def index\_multiplication(t1, t2):\newline
\hspace*{1em}return tuple(tuple(i * j for i, j in zip(t1[i], t2[i])) for i in range(len(t1)))\newline
$<$/answer$>$} \\
\bottomrule
\end{tabular}
\end{table}
\noindent\textbf{Observation.} The base model writes a standard loop that fails test constraints, while the GRPO model uses a concise and correct generator approach.

\begin{table}[H]
\centering
\caption{MBPP Task 72 sample comparing DeepSeek-R1-Qwen-1.5B base and GRPO models. }
\label{tab:mbpp_deepseek_qwen}
\small
\renewcommand{\arraystretch}{1.2}
\begin{tabular}{p{0.46\linewidth} p{0.46\linewidth}}
\toprule
\multicolumn{2}{p{0.94\linewidth}}{\textbf{Question.} Write a python function to check whether the given number can be represented as difference of two squares or not.}\\
\midrule
\textbf{DeepSeek-R1-Qwen-1.5B Base} \xmark &
\textbf{DeepSeek-R1-Qwen-1.5B GRPO} \cmark \\
Pass Rate: \textbf{0/10} &
Pass Rate: \textbf{8/10} \\
\midrule
{\scriptsize\ttfamily
Okay, I need to write a Python function called dif\_Square that checks whether a given number can be represented as difference of two squares or not. Let's think about how to approach this.\newline
\newline
First, I should understand what it means for a number to be expressed as the difference of two squares. The difference of two squares can be written as a$^2$ - b$^2$...\newline
\newline
[... thinking overflow truncated for brevity ...]\newline
\newline
Ah, because in the code, D must be less than S. In this case, D=1, S=1. So, D is not less than S.}\newline &
{\scriptsize\ttfamily
Alright, I need to write a Python function called dif\_Square that checks if a given number can be represented as the difference of two squares. The function should return True if it can, and False otherwise. Let's think about how to approach this.\newline
\newline
[... thinking overflow truncated for brevity ...]\newline
\newline
\#\#\# Solution Code:\newline
```python\newline
def dif\_Square(n):\newline
\hspace*{1em}if n $<$= 0:\newline
\hspace*{2em}return False\newline
\hspace*{1em}for x in range(1, int(n**0.5) + 1):\newline
\hspace*{2em}if n \% x == 0:\newline
\hspace*{3em}y = n // x\newline
\hspace*{3em}if (x \% 2) == (y \% 2):\newline
\hspace*{4em}return True\newline
\hspace*{1em}return False\newline
```} \\
\bottomrule
\end{tabular}
\end{table}

\noindent\textbf{Observation.} The base model exhibits thinking overflow via repetition and fails to produce code. The GRPO checkpoint produces structured reasoning and an executable solution.
\subsubsection{APPS-Intro}
\label{apps:apps_example}

\begin{table}[H]
\centering
\caption{APPS-Intro Task 4524 sample comparing Nemotron base and GRPO models.}
\label{tab:apps_nemotron}
\small
\renewcommand{\arraystretch}{1.2}
\begin{tabular}{p{0.46\linewidth} p{0.46\linewidth}}
\toprule
\multicolumn{2}{p{0.94\linewidth}}{\textbf{Question.} You are given two huge binary integer numbers $a$ and $b$ of lengths $n$ and $m$ respectively. You will repeat the following process: if $b > 0$, then add to the answer the value $a~ \&~ b$ and divide $b$ by $2$ rounding down, and repeat the process again... Your task is to calculate the answer modulo $998244353$.}\\
\midrule
\textbf{Nemotron Base} \xmark &
\textbf{Nemotron GRPO} \cmark \\
Pass Rate: \textbf{0/10} &
Pass Rate: \textbf{7/10} \\
\midrule
{\scriptsize\ttfamily
$<$reasoning$>$\newline
We can solve this problem using a while loop. We iterate through the binary numbers $a$ and $b$ from right to left, performing a bitwise AND operation on them and adding the result to the answer...\newline
$<$/reasoning$>$\newline
$<$answer$>$\newline
```python\newline
def add\_binary(a, b):\newline
\hspace*{1em}n = len(a)\newline
\hspace*{1em}m = len(b)\newline
\hspace*{1em}ans = 0\newline
\hspace*{1em}while m $>$ 0:\newline
\hspace*{2em}ans = (ans $<$$<$ 1) + (a[-1] - '0') \& (b[-1] - '0')\newline
\hspace*{2em}a = a[:-1]\newline
\hspace*{2em}b = b[:-1]\newline
\hspace*{2em}m -= 1\newline
\hspace*{1em}return ans \% 998244353\newline
\newline
n, m = map(int, input().split())\newline
a = input()\newline
b = input()\newline
print(add\_binary(a, b))\newline
```\newline
$<$/answer$>$} &
{\scriptsize\ttfamily
$<$reasoning$>$\newline
The problem can be solved by repeatedly computing the bitwise AND of $a$ and $b$, adding the result to the answer, and then dividing $b$ by $2$ and rounding down. We need to ensure that the result is added modulo $998244353$.\newline
$<$/reasoning$>$\newline
$<$answer$>$\newline
```python\newline
def add\_and\_divide(a, b):\newline
\hspace*{1em}ans = 0\newline
\hspace*{1em}while b $>$ 0:\newline
\hspace*{2em}ans += (a \& b) \% 998244353\newline
\hspace*{2em}b //= 2\newline
\hspace*{1em}return ans \% 998244353\newline
\newline
n, m = map(int, input().split())\newline
a = int(input(), 2)\newline
b = int(input(), 2)\newline
print(add\_and\_divide(a, b))\newline
```\newline
$<$/answer$>$} \\
\bottomrule
\end{tabular}
\end{table}
\noindent\textbf{Observation.} The base model struggles while GRPO cleanly executes the bitwise operations and modulo requirements.
\begin{table}[H]
\centering
\caption{APPS-Intro Task 4176 sample comparing Deepseek 7B base and GRPO models.}
\label{tab:apps_deepseek7b}
\small
\renewcommand{\arraystretch}{1.2}
\begin{tabular}{p{0.46\linewidth} p{0.46\linewidth}}
\toprule
\multicolumn{2}{p{0.94\linewidth}}{\textbf{Question.} Takahashi is organizing a party. At the party, each guest will receive one or more snack pieces. Takahashi predicts that the number of guests at this party will be A or B. Find the minimum number of pieces that can be evenly distributed to the guests in both of the cases predicted.}\\
\midrule
\textbf{Deepseek 7B Base} \xmark &
\textbf{Deepseek 7B GRPO} \cmark \\
Pass Rate: \textbf{0/10} &
Pass Rate: \textbf{9/10} \\
\midrule
{\scriptsize\ttfamily
$<$reasoning$>$\newline
To minimize the number of snack pieces, we need to distribute them in a way that maximizes the number of guests they can be evenly distributed to. In the worst-case scenario, we need to distribute the pieces to both A guests and B guests. Therefore, we need to find the least common multiple (LCM) of A and B...\newline
$<$/reasoning$>$\newline
$<$answer$>$\newline
def main():\newline
\hspace*{1em}A, B = map(int, input().split())\newline
\hspace*{1em}lcm = abs(A*B) // math.gcd(A, B)\newline
\hspace*{1em}print(lcm)\newline
\newline
main()\newline
$<$/answer$>$} &
{\scriptsize\ttfamily
$<$reasoning$>$\newline
We need to find the smallest cost of a route that takes not longer than time T. We can sort the routes by time and then select the one with the smallest cost.\newline
$<$/reasoning$>$\newline
$<$answer$>$\newline
def main():\newline
\hspace*{1em}n, t = map(int, input().split())\newline
\hspace*{1em}routes = sorted([list(map(int, input().split())) for \_ in range(n)], key=lambda x: x[1])\newline
\hspace*{1em}min\_cost = t\newline
\hspace*{1em}for c, t in routes:\newline
\hspace*{2em}if t $<$= t:\newline
\hspace*{3em}min\_cost = min(min\_cost, c)\newline
\hspace*{1em}if min\_cost == t:\newline
\hspace*{2em}print('TLE')\newline
\hspace*{1em}else:\newline
\hspace*{2em}print(min\_cost)\newline
\newline
main()\newline
$<$/answer$>$} \\
\bottomrule
\end{tabular}
\end{table}
\noindent\textbf{Observation.} The base model utilizes an incorrect LCM logic, whereas GRPO resolves the task successfully.
\begin{table}[H]
\centering
\caption{APPS-Intro Task 4288 sample comparing DeepSeek-R1-Qwen-1.5B base and GRPO models.}
\label{tab:apps_deepseek_qwen}
\small
\renewcommand{\arraystretch}{1.2}
\begin{tabular}{p{0.46\linewidth} p{0.46\linewidth}}
\toprule
\multicolumn{2}{p{0.94\linewidth}}{\textbf{Question.} A triple of numbers is said to be poor when two of those numbers are equal but the other number is different from those two numbers. You will be given three integers A, B, and C. If this triple is poor, print Yes; otherwise, print No.}\\
\midrule
\textbf{DeepSeek-R1-Qwen-1.5B Base} \xmark &
\textbf{DeepSeek-R1-Qwen-1.5B GRPO} \cmark \\
Pass Rate: \textbf{0/10} &
Pass Rate: \textbf{9/10} \\
\midrule
{\scriptsize\ttfamily
Okay, I need to solve this problem where I have to determine if a triple of numbers is "poor." Let me read the problem statement again to make sure I understand it correctly.\newline
\newline
So, a triple is considered poor if two of the numbers are equal, but the third is different. For example...\newline
\newline
[... thinking overflow truncated for brevity ...]\newline
\newline
Another test case: 5,5,5,5,5,5,5,5,5,5 $\rightarrow$ len is 1 $\rightarrow$ no.\newline
\newline
Another test case: 5,5,5,5,5,5,5,5,5,5,5 $\rightarrow$ len is 1 $\rightarrow$ no.\newline} &
{\scriptsize\ttfamily
Okay, I need to solve this problem where I have to determine if a given triple of numbers is 'poor'. Let's see what the problem says.\newline\newline So, a triple is considered poor if exactly two of the numbers are equal, and the third is different. That means, for example, if the numbers are 5,7,5, then two are 5 and one is 7. So it's poor. But if all three are the same, like 5,5,5, then it's not poor.\newline
\newline
[... thinking overflow truncated for brevity ...]\newline
\newline
\#\#\# Solution Code\newline
```python\newline
from collections import Counter\newline
\newline
\# Read the input\newline
a, b, c = map(int, input().split())\newline
\newline
\# Create a frequency counter\newline
freq = Counter([a, b, c])\newline
\newline
\# Determine the maximum frequency and the sum of the remaining frequencies\newline
max\_freq = max(freq.values())\newline
sum\_remaining = sum(freq.values()) - max\_freq\newline
\newline
\# Check if the triple is poor\newline
if max\_freq == 2 and sum\_remaining == 1:\newline
\hspace*{1em}print("Yes")\newline
else:\newline
\hspace*{1em}print("No")\newline
```}\\
\bottomrule
\end{tabular}
\end{table}
\noindent\textbf{Observation.}  The base model overthinks until its context window limit is reached without producing executable code, while the GRPO model outputs structured rationale and a working Python solution using Counter.

% \FloatBarrier
% \subsection{Layer and Module Wise Analysis}
% \input{sections/appendix_components/logit_shift}

% \FloatBarrier
% \subsection{Layer and Module Wise Analysis}
% \input{sections/appendix_components/module_wise_weight_update}
% \FloatBarrier

\subsection{Societal Impact}

\paragraph{Positive impact} By optimizing 1.5B–7B parameter models on consumer-grade or single-node hardware (8×A100), this work enables universities, startups, and researchers in resource-constrained environments to develop high-performing reasoning models without needing massive GPU clusters. SLMs require significantly less power for both training and inference. Advancing GRPO dynamics helps reduce the carbon footprint of AI by proving that "smaller" models can achieve "larger" model reasoning capabilities through smarter alignment. 

Efficient SLMs are also ideal for edge applications. This allows for sophisticated reasoning (coding, math, logic) to happen locally on personal devices (phones/laptops), protecting user privacy by removing the need to send data to the cloud.
    
\paragraph{Negative impacts} Lowering the hardware requirements for fine-tuning reasoning models makes it cheaper for malicious actors to create specialized agents for sophisticated phishing, automated exploit generation, or the mass production of convincing misinformation. The study highlights how easily models can fall into "reward hacking" or "mode collapse." If these models are deployed in agentic workflows (e.g., automated financial or legal reasoning) without addressing these instabilities, they may produce confidently incorrect or biased results that are difficult for humans to audit.

% \section{Technical appendices and supplementary material}
% Technical appendices with additional results, figures, graphs, and proofs may be submitted with the paper submission before the full submission deadline (see above). You can upload a ZIP file for videos or code, but do not upload a separate PDF file for the appendix. There is no page limit for the technical appendices. 

% Note: Think of the appendix as ``optional reading'' for reviewers. The paper must be able to stand alone without the appendix; for example, adding critical experiments that support the main claims to an appendix is inappropriate. 

% %%%%%%%%%%%%%%%%%%%%%%%%%%%%%%%%%%%%%%%%%%%%%%%%%%%%%%%%%%%%

% \newpage
% \input{checklist.tex}

\end{document}